\documentclass[10pt]{article} % For LaTeX2e

\usepackage[preprint]{rlj} % Should be uncommented for preprint versions

\usepackage{amssymb}            % Defines common symbols like \mathbb R
\usepackage{mathtools}          % Extends amsmath, providing common math tools
\usepackage{mathrsfs}           % Enables \mathscr, which can work in cases that \mathcal does not
\usepackage{graphicx}           % For including images
\usepackage{subcaption}         % Allows for the use of subfigures and subcaptions
\usepackage[space]{grffile}     % For spaces in image names
\usepackage{url}                % For displaying URLs
\usepackage{lipsum}             % For placeholder text
\usepackage{booktabs}
\usepackage[ruled,vlined]{algorithm2e}

\newcommand{\floor}[1]{\left\lfloor #1 \right\rfloor}

\title{Multi-Resolution Skills for HRL Agents}

\setrunningtitle{Multi-Resolution Skills}

\author{Shashank Sharma, Janina Hoffmann, Vinay Namboodiri}

\emails{ss3966@bath.ac.uk, \ \{jah253,vpn22\}@bath.ac.uk}

\affiliations{
$^{1}$\textbf{Department of Computer Science, University of Bath, UK}\\
$^{2}$\textbf{Department of Psychology, University of Bath, UK}\\
\par % If including additional comments like below, use \par to add some whitespace. 
}

\contribution{
    We characterize a precision-smoothness tradeoff in subgoal-based HRL and show that unconstrained and geometrically unstructured goal representations prevent precise local subgoal selection, contributing to the known performance gap with monolithic agents on agility-demanding tasks. We further show that the optimal subgoal distance is both task- and state-dependent, motivating adaptive multi-resolution selection.
    }
    {
    This characterization applies specifically to subgoal-based HRL methods and does not generalize without modification to options-based or skill-discovery approaches.
    }

\contribution{
    We propose Multi-Resolution Skills (MRS), which learns multiple goal-prediction modules each specialized to a fixed temporal horizon, with a shared backbone to minimize parameter overhead and a jointly trained meta-controller that selects among them based on current state, all within a single end-to-end training phase.
    }
    {
    MRS is implemented on top of the Director architecture; changes are confined to the goal representation and manager policy.
    }

\contribution{
    MRS consistently outperforms fixed-resolution baselines across DeepMind Control Suite, Gym-Robotics, and AntMaze tasks, significantly reduces the performance gap with non-HRL state-of-the-art on agility-demanding benchmarks, and ablations confirm the gains arise from temporal partitioning as an inductive bias.
    }
    {
    None
    }

\keywords{Hierarchical Reinforcement Learning, Reinforcement Learning, Subgoal Discovery, Self-Exploration.} % Your keywords

\summary{
HRL agents decompose the policy into a manager and a worker, enabling long-horizon planning but introducing a structural bottleneck at the manager-worker interface: the manager must propose subgoals that the worker conditions on.
In subgoal-based HRL, these goal representations are typically learned without constraints on reachability or temporal distance — the manager can, in principle, propose any state as a subgoal regardless of whether the worker can reach it.
This undifferentiated representation dilutes goal space capacity across the full state manifold, leaving the manager unable to reliably select precise local subgoals when the task demands it.
We further identify that no single subgoal distance is universally optimal: nearby subgoals enable precise local control but amplify prediction noise, while distant subgoals produce smooth trajectories at the cost of geometric precision at turns and transitions.
The appropriate distance varies by task and by state within a task.

We propose Multi-Resolution Skills (MRS), an HRL framework that addresses both problems by learning multiple goal-prediction modules, each specialized to predict reachable states at a fixed temporal horizon.
Modules share a common backbone with resolution-specific heads, keeping parameter overhead minimal, ensuring performance gains are structural rather than capacity-driven.
A learned meta-controller selects among resolution-specific skill policies based on the current state, enabling dynamic interleaving of fine- and coarse-grained subgoals within a single episode.
All components are trained jointly in a single end-to-end phase with no task-specific tuning.
Evaluated on DeepMind Control Suite, Gym-Robotics, and long-horizon AntMaze tasks, MRS consistently outperforms fixed-resolution baselines and significantly reduces the performance gap between HRL and non-HRL state-of-the-art on agility-demanding benchmarks.
Ablations confirm that the gain arises from temporal partitioning as an inductive bias.
}

\begin{document}

\makeCover  % Create the cover page
\maketitle  % Make the title section

\begin{abstract}
Hierarchical reinforcement learning (HRL) decomposes the policy into a manager and a worker, enabling long-horizon planning but introducing a performance gap on tasks requiring agility.
We identify a root cause: in subgoal-based HRL, the manager's goal representation is typically learned without constraints on reachability or temporal distance from the current state, preventing precise local subgoal selection.
We further show that the optimal subgoal distance is both task- and state-dependent: nearby subgoals enable precise control but amplify prediction noise, while distant subgoals produce smoother motion at the cost of geometric precision.
We propose Multi-Resolution Skills (MRS), which learns multiple goal-prediction modules each specialized to a fixed temporal horizon, with a jointly trained meta-controller that selects among them based on the current state.
MRS consistently outperforms fixed-resolution baselines and significantly reduces the performance gap between HRL and non-HRL state-of-the-art on DeepMind Control Suite, Gym-Robotics, and long-horizon AntMaze tasks.
[Project page: \url{https://sites.google.com/view/multi-res-skills/home}]
\end{abstract}

\vspace{-1.em}

\section{Introduction}
\label{sec:intro}

\vspace{-.5em}

Hierarchical reinforcement learning (HRL) decomposes the policy into a manager that proposes temporally abstract subgoals and a worker that executes primitive actions to reach them \citep{sutton1999between,dayan1992feudal,vezhnevets2017feudal}.
This decomposition enables effective long-horizon planning by reducing the effective decision horizon for each component \citep{hafner2022deep, nachum2018data}.
However, it also introduces a structural information bottleneck at the manager-worker interface: the manager must compress its intent into a finite-dimensional goal signal, and the worker can only condition on that compressed representation.
While this bottleneck is acceptable for long-horizon tasks where coarse guidance suffices, it limits the precision available to the worker on agility-demanding tasks like dense-reward locomotion and manipulation benchmarks where precise, reactive control is essential.
As a result, HRL agents consistently underperform monolithic agents on such tasks \citep{hafner2022deep, hafner2023mastering}, which can hinder practical adoption in settings that require both long-horizon competence and fine-grained control.

One source of this gap, in subgoal-based HRL, is the structure of the learned goal representation.
The manager proposes subgoals from a goal space that is typically learned without constraints on reachability or temporal distance from the current state.
Thus, the manager can, in principle, propose any state as a subgoal, regardless of whether the worker can reach it \citep{nachum2018data,hafner2022deep,vezhnevets2017feudal}.
This unconstrained goal space has two consequences: the worker is frequently directed toward unreachable targets \citep{zhang2020generating}, and representational capacity is diluted across the full state manifold rather than concentrated on the locally reachable states relevant for precise control \citep{nachum2018near}.
Together, these compound the information bottleneck and contribute to the precision deficit observed in practice.

\begin{figure}
    \centering
    \begin{subfigure}[b]{0.9\textwidth}
        \centering
        \includegraphics[width=\textwidth]{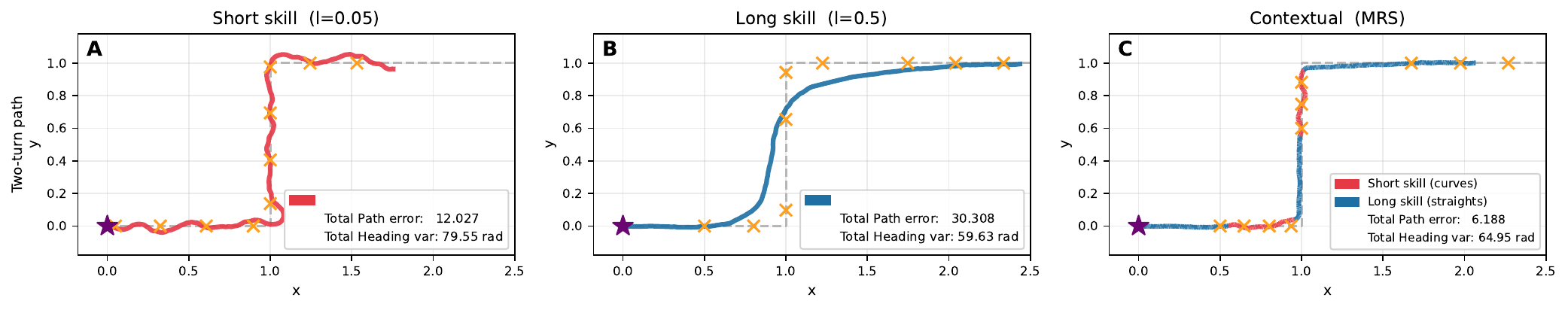}
    \end{subfigure}
    \hfill
    \begin{subfigure}[b]{0.9\textwidth}
        \centering
        \includegraphics[width=\textwidth]{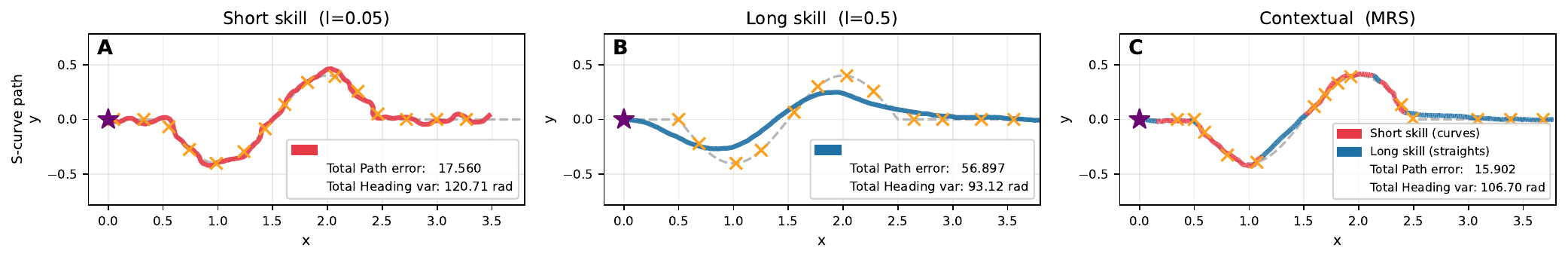}
    \end{subfigure}
    \caption{A point agent navigates two paths: a two-turn path (top) and an S-curve (bottom), by tracking subgoals sampled from future positions on the path with fixed additive noise. The short-skill agent (\textcolor{red}{red}) uses nearby subgoals: it follows curves precisely but accumulates jitter as the same absolute prediction noise produces large angular deviations when the target is close. The long-skill agent (\textcolor{blue}{blue}) uses distant subgoals: it moves smoothly on straights but cuts corners and accumulates geometric error at transitions, as a fixed subgoal cannot capture the rapid heading change required at turns. The contextual agent (MRS, \textcolor{red}{red}/\textcolor{blue}{blue}) selects short skills near turns and long skills on straights: it achieves lower total path error than either fixed alternative on the two-turn path, and correctly defaults to short skills throughout the continuously curved S-path. Crosses mark the assigned subgoal positions. Total path error and heading variance are reported in the legend. This illustrates that the optimal subgoal distance is both task- and state-dependent, and that no fixed skill length is universally optimal, motivating the adaptive multi-resolution selection in MRS. Note that the switching rule here is hand-coded based on local curvature.
    % in MRS, the meta-controller learns this selection from data.
    }
    \label{fig:toy_sim}
    \vspace{-1.5em}
\end{figure}

A related but distinct difficulty concerns not the structure of the goal space but the distance of the proposed subgoal.
Even if the goal space were appropriately constrained, the optimal subgoal distance is neither fixed nor task-invariant.
We illustrate this with a simulation of a point agent navigating paths with straight segments and sharp turns (Fig. \ref{fig:toy_sim}).
When the subgoal is nearby, the agent follows curves precisely but is sensitive to prediction noise, where small errors produce large angular deviations.
When the subgoal is distant, the agent moves smoothly but accumulates geometric error at turns.
Neither strategy dominates across all path geometries.
An agent that selects short skills near turns and long skills on straights outperforms either fixed alternative on heterogeneous paths, establishing that the optimal subgoal distance is both task- and state-dependent.

Motivated by these observations, we propose \emph{Multi-Resolution Skills} (MRS), an HRL framework that addresses both problems.
Rather than learning a single unconstrained goal representation, MRS learns multiple goal-prediction modules, each specialized to predict reachable states at a fixed temporal horizon.
Because each module is conditioned on the current state and constrained to its temporal scale, its goal space is compact and concentrated on locally reachable states, and the manager can select precise local subgoals through short-horizon modules and smooth long-range targets through long-horizon modules.
All modules share a common backbone and differ only in their resolution-specific heads, so performance gains are structural rather than capacity-driven.
A learned meta-controller selects among resolution-specific skill policies based on the current state, enabling dynamic interleaving of fine- and coarse-grained behavior.
The goal-prediction modules, skill policies, and meta-controller are jointly trained in a single end-to-end phase, with no task-specific tuning required.

MRS relates to two adjacent lines of prior work, discussed in detail in Sec. \ref{sec:related}.
Unsupervised skill discovery methods \citep{eysenbach2018diversity, sharma2020dynamics, jiang2022unsupervised} learn diverse behaviors without external rewards but impose no temporal structure on the skill space and cannot guarantee coverage of all local state transitions.
Goal-conditioned HRL methods \citep{nachum2018data, hafner2022deep,vezhnevets2017feudal, zhang2020generating} use subgoal prediction but learn unconstrained goal representations.
MRS retains the goal-conditioned framework but adds explicit temporal structure to the goal representation.
HiPPO \citep{li2019sub} also addresses fixed temporal abstraction but randomizes skill length at each manager step for robustness, treating it as a fixed stochastic process rather than a learned, state-dependent decision.
We implement MRS on top of the Director \citep{hafner2022deep} architecture, modifying only the goal representation and manager policy.

Our contributions are as follows:

\begin{enumerate}

\item We characterize a precision-smoothness tradeoff in subgoal-based HRL and show that unconstrained goal representations contribute to the known performance gap with monolithic agents on agility-demanding tasks. We further show that the optimal subgoal distance is both task- and state-dependent, motivating adaptive multi-resolution selection. (Fig. \ref{fig:toy_sim}, Fig. \ref{fig:choice_hist}, Fig. \ref{fig:choice_tsne})

\item We propose MRS, which learns multiple goal-prediction modules each specialized to a fixed temporal horizon, with a shared backbone to minimize parameter overhead and a jointly trained meta-controller that selects among them based on current state, all within a single end-to-end training phase. (Sec. \ref{sec:methodology})

\item We empirically show that MRS consistently outperforms fixed-resolution baselines (Tab. \ref{tab:ablation_single}), significantly reduces the performance gap with non-HRL state-of-the-art on agility-demanding benchmarks (Tab. \ref{tab:main_results}), and ablation studies confirm that the gains arise from temporal partitioning as an inductive bias. (Sec. \ref{sec:results})

\end{enumerate}

\vspace{-.5em}

\section{Preliminaries}
\label{sec:background}

\textbf{World Model.}
We build on the Recurrent State Space Model (RSSM) \citep{hafner2019learning}, which encodes a state representation $s_t$ using the observations $o_{t}$ and previous state-actions $(s_{t-1}, a_{t-1})$ via a learned representation model $\texttt{repr}(s_t|o_t,s_{t-1}, a_{t-1})$.
A dynamics model predicts future states $\texttt{dyn}(s_{t+1}|s_t,a_t)$ without requiring observations, enabling the agent to imagine trajectories entirely within the learned latent space.
All policies in Director and MRS operate on $s_t$ as their state representation.

\textbf{Manager-Worker Decomposition.}
Director \citep{hafner2022deep} introduces a two-level hierarchy on top of the RSSM.
The manager $\pi_M$ proposes a subgoal $s_g$ that updates every $K$ steps.
The worker $\pi_W$ receives $(s_t, s_g)$ at each step and outputs a primitive action $a_t$, trained with an intrinsic reward $R^W_t = \texttt{cosine\_max}(s_t, s_g)$ that encourages it to move towards the subgoal.
The manager is trained to maximize the sum of extrinsic task rewards and an exploratory bonus over each $K$-step interval; we describe the exploratory objective in Sec.~\ref{sec:training}.

\textbf{Goal Autoencoder.}
Directly predicting subgoals in the full RSSM state space poses a high-dimensional continuous-control problem for the manager.
The Director therefore learns a goal autoencoder with encoder $\text{Enc}_\phi(z \mid s_t)$ and decoder $\text{Dec}_\phi(z) \approx s_t$, which compresses world model states into a compact categorical latent $z$ using a mixture of categoricals as the prior.
The manager operates in this latent space, sampling $z \sim \pi_M(z \mid s_t)$ and decoding to obtain $s_g = \text{Dec}_\phi(z)$.
The autoencoder is trained on all states encountered in the replay buffer, without constraints on reachability or temporal distance from the current state, the limitation that MRS addresses.

\textbf{Policy Optimization.}
Both, manager and worker, are trained as actor-critics optimized via policy gradients on trajectories imagined using the RSSM, using $\lambda$-returns for value estimation and entropy regularization to encourage exploration.
MRS uses the same optimization procedure with a modified manager objective (Sec. \ref{sec:training}).

\vspace{-.5em}

\section{Methodology}
\label{sec:methodology}

\vspace{-.25em}

\subsection{Skills as Constrained State Transitions}
\label{sec:skills}

\begin{figure}
    \centering
    \begin{subfigure}[b]{0.45\textwidth}
        \centering
        \includegraphics[width=\textwidth]{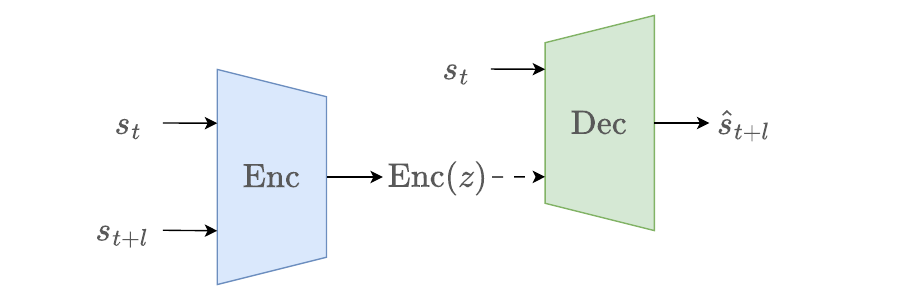}
        \caption{Skill CVAE architecture}
        \label{fig:sk_cvae}
    \end{subfigure}
    \hfill
    \begin{subfigure}[b]{0.45\textwidth}
        \centering
        \includegraphics[width=\textwidth]{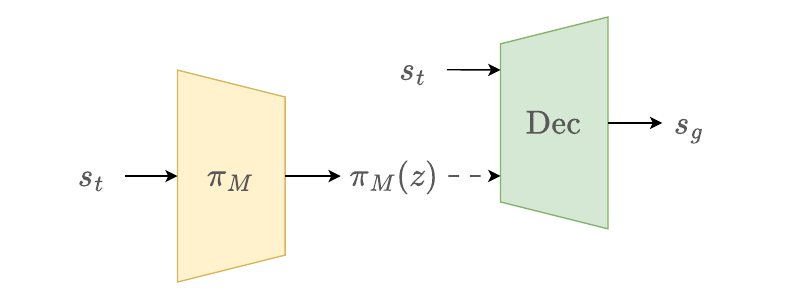}
        \caption{Acting using Skill CVAE}
        \label{fig:sk_actor}
    \end{subfigure}
    \caption{Illustrations of the abstract state transition-based control for the manager. \textit{Dashed} arrows indicate sample propagation from the predicted distribution. \textbf{(a)} Skill CVAE, where the Encoder encodes initial and final states $(s_t,s_{t+l})$ to a latent skill space and the Decoder reconstructs the final state using the initial state $s_t$ and a sampled skill variable $z$. \textbf{(b)} The manager predicts the latent skills and then uses the Decoder to generate goals for the worker.}
    \label{fig:sk_arch}
    \vspace{-1.em}
\end{figure}

The Director's goal autoencoder learns to compress and reconstruct states unconditionally, without reference to the current state $s_t$.
As a result, the manager can decode any point in the latent space into a subgoal, regardless of whether it is reachable from $s_t$ within the worker's capabilities.
We replace the goal autoencoder with a Conditional VAE (CVAE) that learns to model state transitions of a fixed length $l$.
Given the current state $s_t$, the CVAE learns to encode and decode future states $s_{t+l}$ that are reachable in exactly $l$ steps under the current policy.
Training pairs $(s_t, s_{t+l})$ are extracted from the replay buffer by taking states $l$ steps apart within each trajectory.
The CVAE encoder $\text{Enc}_\phi(z \mid s_t, s_{t+l})$ maps the transition to a latent skill variable $z$, and the decoder $\text{Dec}_\phi(s_t, z)$ reconstructs the target state conditioned on both $s_t$ and $z$ (Fig. \ref{fig:sk_cvae}).
Conditioning the decoder on $s_t$ ensures that $z$ encodes only the transition rather than the full state, keeping the latent space compact and structured relative to the current state.
Therefore, skills $z$ correspond to abstract state-transitions possible from the current state $s_t$.
The CVAE is trained on the ELBO objective:

\vspace{-1.em}

\begin{equation}
    \label{eq:skvae_loss}
    \mathcal{L}(\phi) = \left\Vert s_{t+l} - \text{Dec}_\phi(s_t, z)\right\Vert^2 + \beta\text{KL}\left[\text{Enc}_\phi(z \mid s_t, s_{t+l}) || p(z)\right], \quad z \sim \text{Enc}_\phi(z \mid s_t, s_{t+l})
\end{equation}

where $p(z)$ is a mixture of $8 \times 8$-dimensional categoricals used as the prior.
At inference, the manager samples $z \sim \pi_{M}(z \mid s_t)$ and decodes to obtain the subgoal $s_g = \text{Dec}_\phi(s_t, z)$, constraining subgoal selection to the neighbourhood of states reachable from $s_t$ in $l$ steps (Fig. \ref{fig:sk_actor}).
A single skill length $l$, however, cannot be universally optimal (Sec.~\ref{sec:intro}); the appropriate subgoal distance varies by task and by state within a task.
This motivates learning skills at multiple temporal resolutions simultaneously.

\subsection{Multi-Resolution Skill Modules}
\label{sec:mrsk}

\begin{figure}
    \centering
    \begin{subfigure}[b]{0.48\textwidth}
        \centering
        \includegraphics[width=\textwidth]{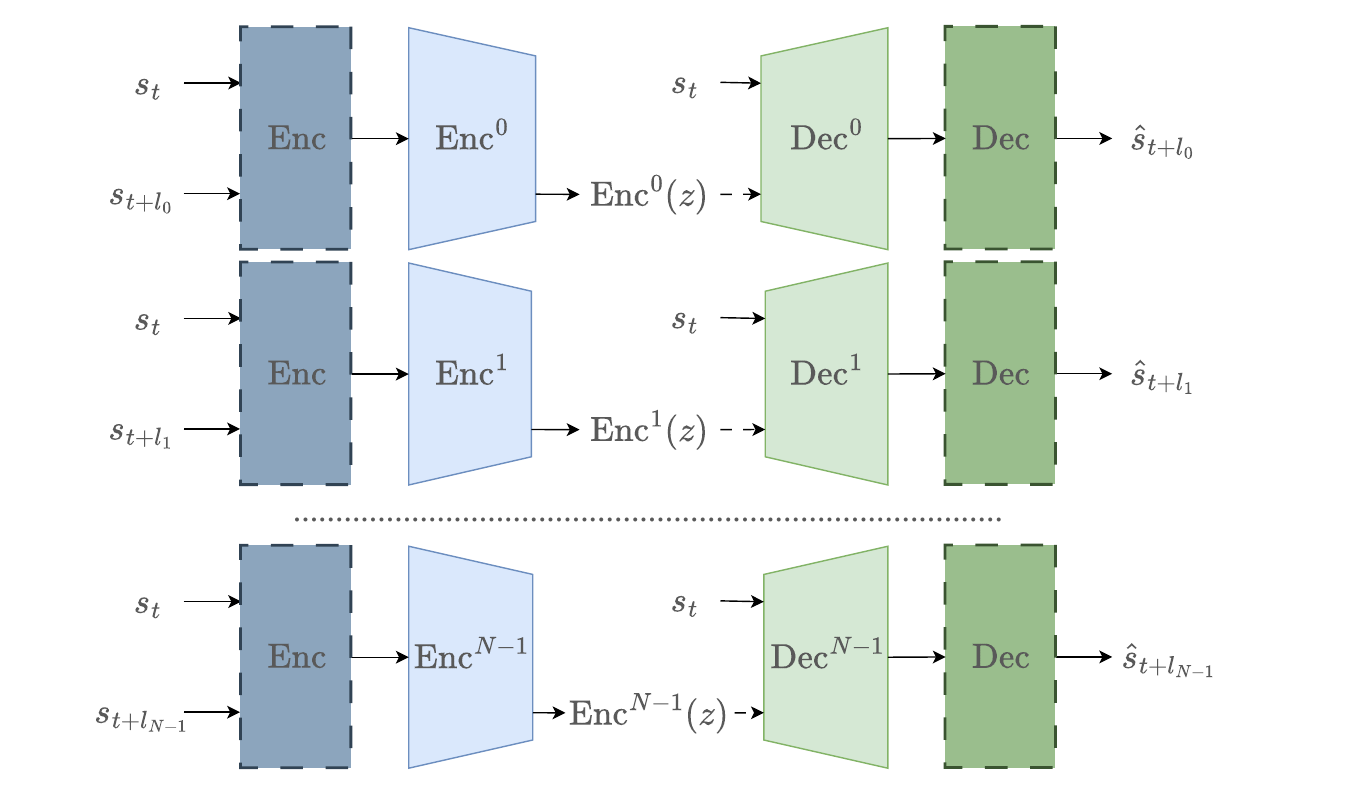}
        \caption{Learning Multi-Resolution Skill CVAEs}
        \label{fig:mrsk_cvae}
    \end{subfigure}
    % \par\bigskip
    \begin{subfigure}[b]{0.48\textwidth}
        \centering
        \includegraphics[width=\textwidth]{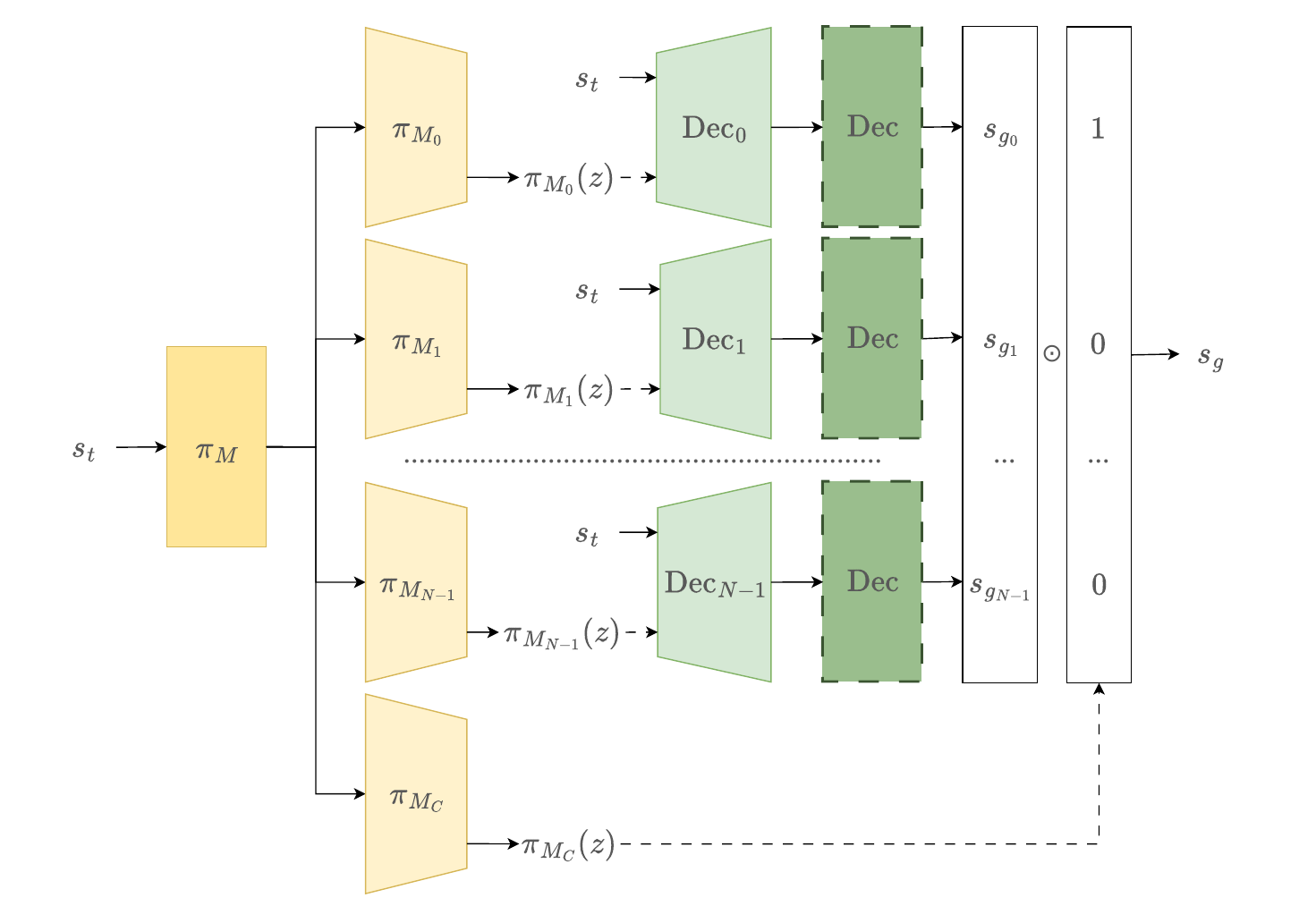}
        \caption{Acting using Multi-Resolution Skill CVAEs}
        \label{fig:mrsk_actor}
    \end{subfigure}
    \caption{Architectures for learning and acting using Multi-Resolution Skills ($l_i \in \{l_0,l_1,...,l_{N-1}\}$). Dashed boundaries indicate shared layers. \textbf{(a)} Separate CVAEs are learned for each temporal resolution $l_i$. The $\text{Enc}$ and $\text{Dec}$ modules represent the common layers of the Encoders and the Decoders, respectively. Each $\text{Enc}_i$ is the resolution-specific encoder output layer, and each $\text{Dec}_i$ is the resolution-specific decoder input layer. \textbf{(b)} The manager's policy has $N+1$ output heads. $N$ skill heads $\pi_{M_i}$ that predict the resolution-specific skill latents and choice head $\pi_{M_C}$ that predicts an $N$-dimensional one-hot distribution. Samples from the skill latents are used to predict sub-goals using the respective Decoders, then the choice sample from $\pi_{M_C}$ selects one of the sub-goals as $s_g$ by gating.}
    \label{fig:mrsk_arch}
    \vspace{-1.em}
\end{figure}

We learn $N$ CVAEs in parallel, indexed by resolution $i$, each specialized to a fixed temporal horizon $l_i$.
We choose $l_i \in \{K, 2K, 4K, 8K\}$, where $K$ is the manager's subgoal refresh interval.
This geometric spacing is principled: $l = K$ is the shortest meaningful skill, corresponding to one full manager interval, and doubling gives logarithmically uniform coverage of the temporal horizon.
% Logarithmic rather than linear spacing is appropriate because the behavioral difference between skill lengths is largest at short horizons, the precision-smoothness tradeoff is most sensitive to changes in $l$ when $l$ is small, so geometric spacing concentrates resolution where it matters most.

% To prevent parameter growth from scaling with $N$, all CVAEs share a common backbone: the encoder layers $\text{Enc}_\phi$ and decoder layers $\text{Dec}_\phi$ are shared, with only the resolution-specific input layer $\text{Enc}^i_\phi$ and output layer $\text{Dec}^i_\phi$ kept separate per resolution.
To prevent parameter growth from scaling with $N$, all CVAEs share a common backbone.
All layers are shared except the encoder's last layer $\text{Enc}_\phi^i$, and the decoder's first layer $\text{Dec}_\phi^i$, which are resolution-specific (Fig. \ref{fig:mrsk_cvae}).
This means the parameter overhead of adding $N$ resolutions is limited to $N$ pairs of lightweight heads rather than $N$ full models.
Training extracts pairs $(s_t, s_{t+l_i})$ at each resolution from the replay buffer and optimizes the summed ELBO:

\vspace{-1.em}

\begin{equation}
    \mathcal{L}(\phi) = \sum_{i=0}^{N-1} \left\Vert s_{t+l_i} - \text{Dec}^i_\phi(s_t, z_i)\right\Vert^2 + \beta \text{KL}\left[\text{Enc}^i_\phi(z_i \mid s_t, s_{t+l_i}) || p(z)\right], \quad z_i \sim \text{Enc}^i_\phi
    \label{eq:mrsk_loss}
\end{equation}

\vspace{-.5em}

Shared layers receive gradients from all $N$ resolutions simultaneously; resolution-specific heads receive gradients only from their own resolution.
All CVAEs are trained jointly with the rest of the agent in a single end-to-end phase.

\paragraph{What about longer skills?}
As skill length $l$ increases, the target state $s_{t+l}$ becomes progressively less predictable from $s_t$ as the agent can take many different trajectories in $l$ steps.
In the limit $l \to \infty$, the target state is statistically independent of the starting state, and the conditional distribution collapses to the marginal over all states.
The $\infty$-skill makes this limiting case explicit: adding an unconditional VAE trained on states from the replay buffer without conditioning on any $s_t$, producing subgoals that are independent of the current state by construction.

Beyond its conceptual role, the $\infty$-skill addresses a practical failure mode in joint skill learning.
The manager policy and the finite CVAEs are co-dependent: the CVAEs learn from trajectories generated by the manager, while the manager learns to use the skills the CVAEs represent, a circular dependency well-known to cause degenerate solutions in joint skill learning \citep{bacon2017option, eysenbach2018diversity}.
If the finite CVAEs saturate early, converging to a small set of skills before the manager's policy has sufficiently explored them, the manager collapses into repeating those transitions, which in turn deprives the CVAEs of diverse training data.
% This chicken-and-egg dynamic can lock the agent into a degenerate solution.
The $\infty$-skill breaks this cycle: because it does not depend on any finite-length transition, it remains well-defined regardless of CVAE training state, allowing the manager to generate diverse subgoals even when the finite CVAEs have not yet converged.
In practice, the agent initially relies on the $\infty$-skill and gradually shifts toward the finite skills as the CVAEs mature, a trend visible in the evolution of the choice distribution during training (Fig.~\ref{fig:choice_hist}).
Results in Sec.~\ref{sec:results} show that MRS remains competitive without the $\infty$-skill on most tasks, confirming that it functions as a robustness mechanism against early CVAE saturation rather than a core architectural requirement.
The full resolution set used in all experiments is therefore $l_i \in \{K, 2K, 4K, 8K, \infty\}$, with $N = 5$ total skill modules including the $\infty$-skill.

\vspace{-.5em}

\subsection{The Meta-Controller}
\label{sec:metacontroller}

The manager policy has $N$ skill heads and one choice head (Fig.~\ref{fig:mrsk_actor}).
Each skill head $\pi_{M_i}(z_i \mid s_t)$ predicts a latent skill distribution for its resolution, and the choice head $\pi_{M_C}(c \mid s_t)$ predicts a one-hot $N$-dimensional distribution over resolutions.
At each manager step, all $N$ skill heads sample latents $z_i \sim \pi_{M_i}(z_i \mid s_t)$ to decode to candidate subgoals (Eq. \ref{eq:mrsk_goal_pred}) and select from those via gating (Eq. \ref{eq:mrsk_goal_select}).

\vspace{-1.em}

\begin{gather}
    \label{eq:mrsk_goal_pred}
    s_g^i = \text{Dec}^i_\phi(s_t, z_i) \\
    \label{eq:mrsk_goal_select}
    s_g = \sum_{i=0}^{N-1} c_i \cdot s_g^i, \quad c \sim \pi_{M_C}(c \mid s_t)
\end{gather}

\vspace{-.5em}

The selected subgoal $s_g$ is passed to the worker and held fixed for $K$ steps.
The consequences of this discrete one-hot gating for policy optimization are discussed in Sec.~\ref{sec:training}.

\subsection{Training Objective}
\label{sec:training}

\textbf{CVAE training.}
The $N$ skill CVAEs are trained jointly on Eq.~\ref{eq:mrsk_loss}, updated from the replay buffer throughout training.

\textbf{Manager and worker.}
Both policies are trained as actor-critics optimized via policy gradients on trajectories imagined through the RSSM, using $\lambda$-returns for value estimation and entropy regularization.
% The manager policy gradient separates cleanly into a choice term and $N$ per-head terms weighted by the choice indicator $c_i$, enabling joint optimization of all heads in a single update.
The manager policy gradient takes the form:

\vspace{-1.em}

\begin{equation}
    \nabla \mathcal{J}(\pi_M) = \mathbb{E}\left[ (\sum_{i=0}^{N-1} c_i \underbrace{\nabla \log \pi_{M_i}(z_i \mid s_t)}_{\text{Skill head } i} + \underbrace{\nabla \log \pi_{M_C}(c \mid s_t)}_{\text{Choice head}}) \cdot A_t \right]
    % \nabla_M J = \mathbb{E}_\tau\Bigg[\sum_{k=0}^{\lfloor T/K\rfloor-1}\Big(\underbrace{\nabla_M\log \pi_{M_C}(c_k\!\mid\!s_{kK})}_{\text{Choice head}} + \sum_{i=0}^{N-1} c_{k,i} \underbrace{\nabla_M\log \pi_{M_i}(z_{k,i}\!\mid\!s_{kK})}_{\text{Skill head } i}\Big) . A_t \Bigg]
    \label{eq:manager_pg}
\end{equation}

\vspace{-.5em}

where $A_t$ is the advantage estimate and $c_i \in \{0,1\}$ is the one-hot choice indicator. Because $c_i = 0$ for all unchosen heads at each step, the gradient with respect to skill head $i$'s parameters is explicitly zeroed when that head is not selected, a direct consequence of the objective formulation rather than of the gating architecture.
Each skill head, therefore, receives a gradient only from manager steps where it was chosen, keeping heads specialized to their own resolution and preventing gradient contamination across resolutions.
Full derivations and loss definitions are provided in Appendix~\ref{app:mrs_policy_grads}.

\textbf{Exploratory reward.}
To encourage the manager to discover the full diversity of state transitions in the environment across all temporal scales.
Starting at state $s_0$, we reward the agent for executing transitions $(s_0,s_t)$ that are not yet well-modeled by any skill module.
This is implemented as the minimum CVAE reconstruction error across all $N$ finite skills, selecting the error from whichever module currently best models the transition, and rewarding the manager for transitions that remain poorly explained by the best-fitting module:

\vspace{-1.em}

\begin{equation}
    R^\text{Expl}_t = \min_i \left\Vert s_t - \text{Dec}^i\phi(s_0, z_i)\right\Vert^2, \quad z_i \sim \text{Enc}^i_\phi(z \mid s_0, s_t)
    \label{eq:expl_rew}
\end{equation}

All components: skill CVAEs, skill policies, choice head, and worker, are optimized jointly throughout a single training phase with no task-specific tuning required.
% See Sec. \ref{sec:app_training_details} for full architectural and training details.

\vspace{-.5em}

\section{Experiments}
\label{sec:results}

\vspace{-.5em}

\subsection{Experimental Setup}
\label{sec:setup}

We evaluate MRS across three benchmark families that test distinct capabilities.
The DeepMind Control Suite (DMC) \citep{tassa2018deepmind} provides six continuous control tasks
(\texttt{walker\_run}, \texttt{quadruped\_run}, \texttt{cheetah\_run}, \texttt{hopper\_hop}, \texttt{cartpole\_swingup}, \texttt{pendulum\_swingup})
with dense rewards, testing whether MRS reduces the agility gap relative to HRL baselines.
Gymnasium-Robotics \citep{gymnasium_robotics2023github} provides two sparse-reward ($0$ at goal, $-1$ otherwise) manipulation tasks (\texttt{fetch\_push}, \texttt{fetch\_pick\_n\_place}) that require precise goal-directed control over short horizons.
AntMaze tests long-horizon ($3000$-step) sparse-reward ($1$ at goal, $0$ otherwise) navigation, in which the agent must plan across extended sequences of subgoals without intermediate feedback.

We compare against three baselines.
Director \citep{hafner2022deep} is the state-of-the-art HRL agent on which MRS is built, and represents the unconstrained subgoal HRL baseline.
DreamerV3 \citep{hafner2023mastering} is the state-of-the-art non-HRL agent, and represents the agility ceiling that HRL has historically failed to match.
HiPPO \citep{li2019sub} is included as a temporal-abstraction baseline that executes subpolicies for random time-lengths, operates on privileged state observations, and uses on-policy RL.

Both MRS and Director are trained every $8$th environment step using self-collected data from the replay buffer.
% Director's default update frequency is every 16th step; we increase it to match the higher gradient frequency required by MRS's $N$ policy heads, ensuring that all performance differences are attributable solely to the proposed architectural changes.
DreamerV3 trains every 2nd step, making it approximately $4\times$ more compute-intensive.
We use skill lengths $L = [64, 32, 16, 8, \infty]$ for all MRS experiments.
All agents are evaluated across $4$ seeds; we report the mean and standard deviation.
Full hyperparameters and architecture details are provided in Appendix~\ref{sec:app_training_details}.

\vspace{-.5em}

\subsection{Main Results}
\label{sec:main_results}

\vspace{-.5em}

\begin{figure}
    \centering
    \begin{subfigure}[b]{0.24\textwidth}
        \centering
        \includegraphics[width=\textwidth]{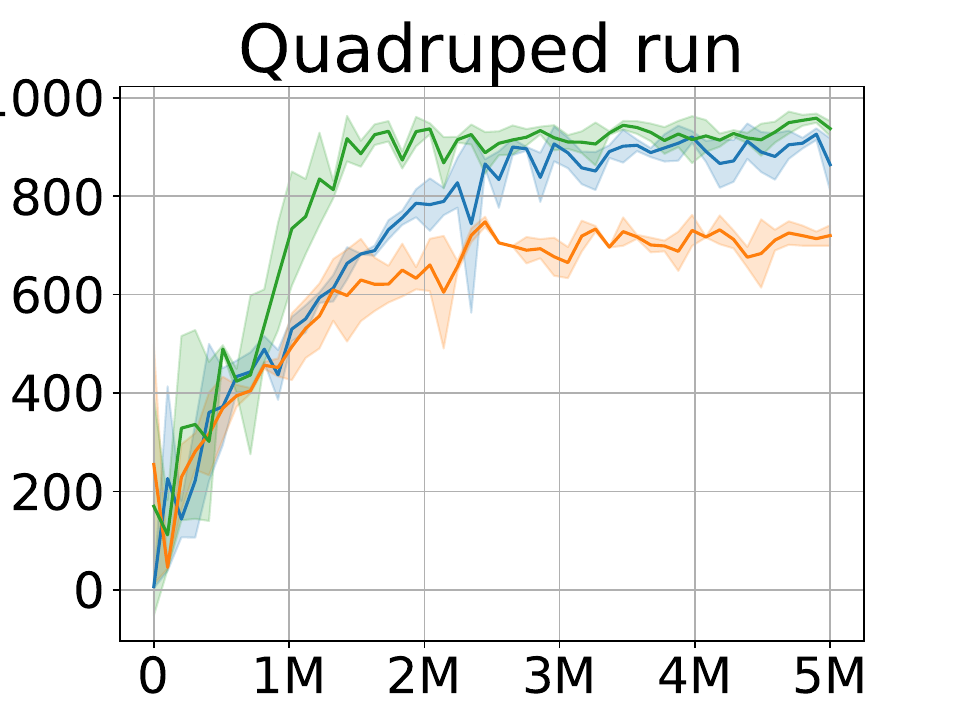}
    \end{subfigure}
    \begin{subfigure}[b]{0.24\textwidth}
        \centering
        \includegraphics[width=\textwidth]{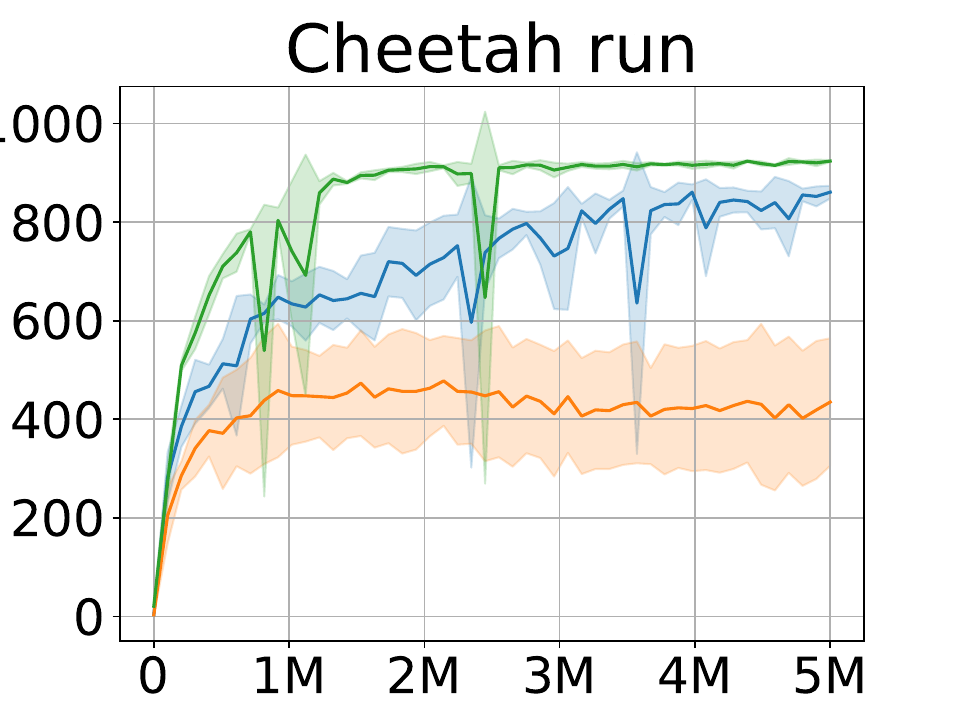}
    \end{subfigure}
    \begin{subfigure}[b]{0.24\textwidth}
        \centering
        \includegraphics[width=\textwidth]{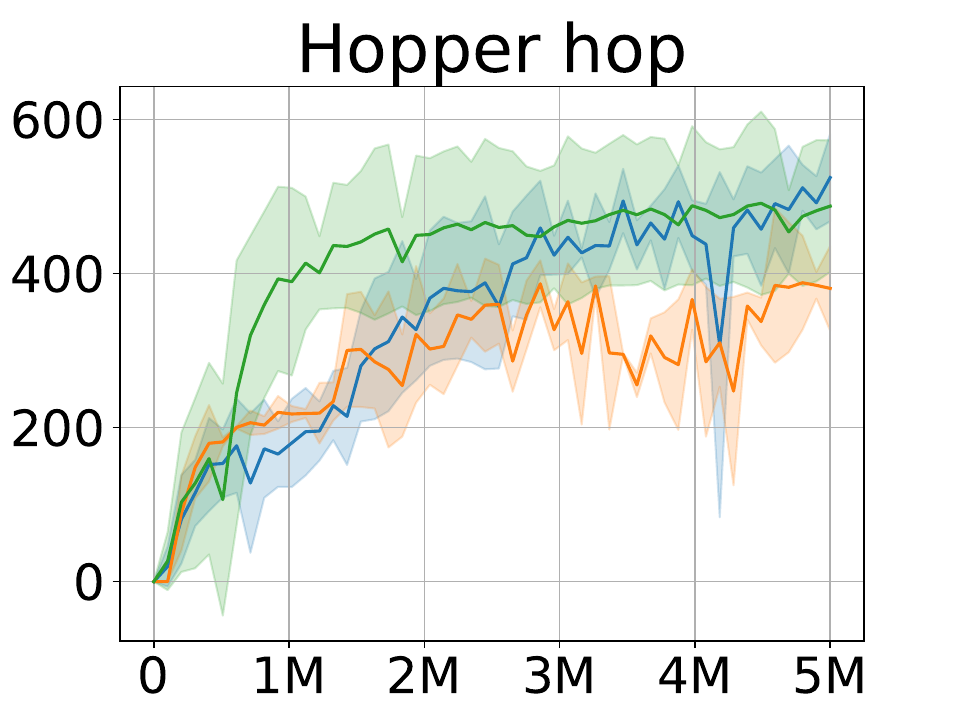}
    \end{subfigure}
    \begin{subfigure}[b]{0.24\textwidth}
        \centering
        \includegraphics[width=\textwidth]{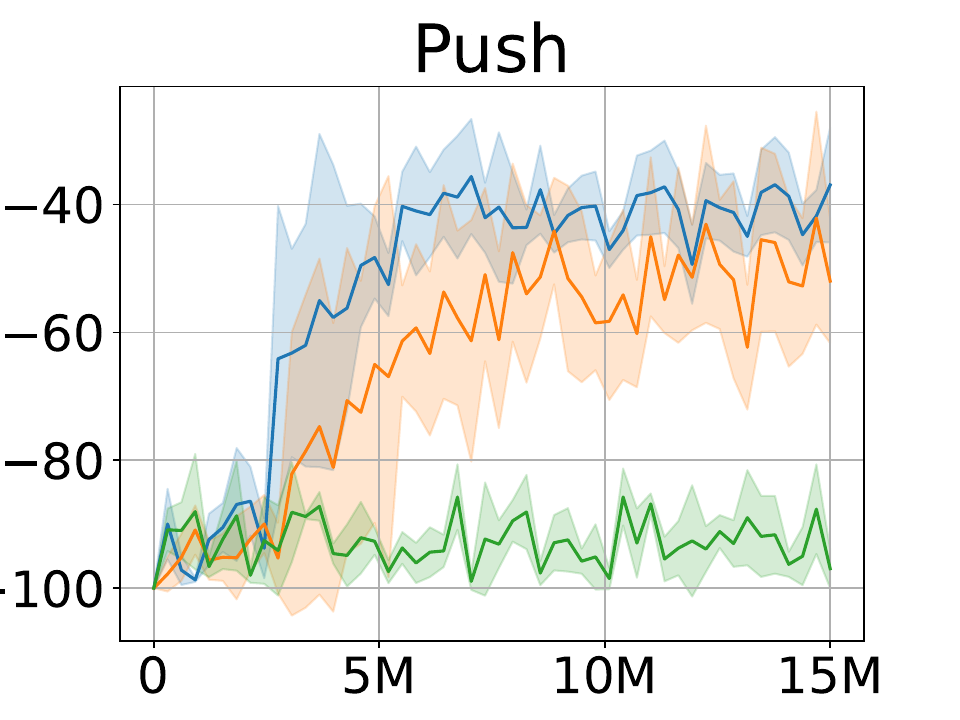}
    \end{subfigure}
    \begin{subfigure}[b]{0.48\textwidth}
        \centering
        \includegraphics[width=\textwidth]{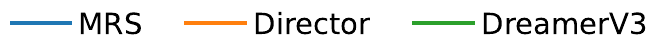}
    \end{subfigure}
    \caption{Episodic scores from MRS (ours), Director, and DreamerV3. The plot shows the total rewards (mean and standard deviation) received in an episode vs the environmental step at the representative DMC (first $3$) and Gym-Robotics (Push) tasks. It can be seen that MRS noticeably improves the base model's performance in all cases while maintaining computational efficiency.}
    \label{fig:main_curves}
    \vspace{-1.em}
\end{figure}

Table~\ref{tab:main_results} reports final performance across all benchmarks.
Fig.~\ref{fig:main_curves} shows learning curves for the representative DMC and Gym-Robotics tasks; full results for all tasks are provided in Appendix~\ref{app:full_train_curves}.

\begin{table}
    \centering
    \begin{tabular}{lcccc}
        \toprule
        \textbf{Task} & \textbf{MRS (ours)} & \textbf{Director} & \textbf{DreamerV3} & \textbf{HiPPO} \\
        \midrule
        \multicolumn{5}{l}{\textit{DeepMind Control Suite (locomotion)}} \\
        \texttt{walker\_run} & $\mathbf{745 \pm 21}$ & $701 \pm 19$ & $\underline{856 \pm 12.24}$ & $472 \pm 85$ \\
        \texttt{quadruped\_run} & $\mathbf{925 \pm 19}$ & $719 \pm 20$ & $\underline{937 \pm 15}$ & $572 \pm 25$ \\
        \texttt{cheetah\_run} & $\mathbf{866 \pm 29}$ & $435 \pm 129$ & $\underline{922 \pm 6}$ & $458 \pm 30$ \\
        \texttt{hopper\_hop} & $\mathbf{\underline{511 \pm 56}}$ & $380 \pm 55$ & $487 \pm 85$ & $102 \pm 37$ \\
        \texttt{cartpole\_swingup} & $842 \pm 6$ & $684 \pm 60$ & $852 \pm 3$ & $\underline{\mathbf{852 \pm 10}}$ \\
        \texttt{pendulum\_swingup} & $526 \pm 297$ & $474 \pm 360$ & $748 \pm 58$ & $\mathbf{\underline{817 \pm 11}}$ \\
        \midrule
        \multicolumn{5}{l}{\textit{Gymnasium-Robotics (manipulation, sparse reward)}} \\
        \texttt{fetch\_push} & $\mathbf{\underline{-36 \pm 8}}$ & $-52 \pm 9$ & $-87 \pm 7$ & $-100 \pm 0$ \\
        \texttt{fetch\_pick\_place} & $\mathbf{\underline{-29 \pm 5}}$ & $-41 \pm 8$ & $-94 \pm 3$ & $-100 \pm 0$ \\
        \midrule
        \multicolumn{5}{l}{\textit{AntMaze (long-horizon, sparse reward)}} \\
        Medium & $\mathbf{\underline{2299 \pm 296}}$ & $2035 \pm 470$ & $0 \pm 0$ & $0 \pm 0$ \\
        Large & $\mathbf{\underline{2224 \pm 395}}$ & $1982 \pm 280$ & $0 \pm 0$ & $0 \pm 0$ \\
        \bottomrule
    \end{tabular}
    \caption{Final performance as cumulative episodic rewards (mean $\pm$ std across $4$ seeds) on DMC, Gym-Robotics, and AntMaze. HiPPO uses ground-truth state observations, whereas all other methods operate on pixels. Best HRL result \textbf{bold}; overall best \underline{underlined}.}
    \label{tab:main_results}
\end{table}

The results reveal a clear pattern across prior methods: Director excels on sparse-reward tasks requiring sustained planning, Gym-Robotics and AntMaze, but consistently underperforms on agility-demanding DMC tasks.
DreamerV3 is competitive or superior on DMC but fails completely on Gym-Robotics, where its flat policy cannot reliably perform in sparse-reward tasks.
Prior work, therefore, forces a choice between long-horizon competence and agility.
MRS achieves competitive performance across all three regimes, matching or exceeding Director on Gym-Robotics and AntMaze, and reducing the gap with DreamerV3 on DMC.
HiPPO performs well on the simple swingup tasks, but underperforms on the complex tasks.
% The \texttt{cheetah\_run} result warrants particular attention.
% MRS scores approximately 850 versus Director's 425 despite the choice histogram for this task (Fig.~\ref{fig:choice_hist}) showing a roughly uniform distribution across all five skill lengths at convergence.
% The t-SNE analysis in Sec.~\ref{sec:qualitative} reveals that skill selection is spatially structured in the latent state space, with distinct state regions showing consistent preferences for particular temporal horizons.
% We interpret the large performance gap as evidence that reliable access to contextually appropriate subgoals in the states where subgoal precision matters most is sufficient to drive substantial performance gains, even when aggregate skill usage appears uniform.
% Director, constrained to a single unconditional goal distribution, cannot vary subgoal precision based on local state context.
On AntMaze, MRS yields gains in both final performance and convergence speed.
We note this as an observed benefit consistent with the higher sample efficiency of constrained subgoal prediction.

\vspace{-.5em}

\subsection{Ablations}
\label{sec:ablations}

\vspace{-.5em}

\textbf{Single-resolution baselines.}
We train four agents each using only a single fixed skill length from scratch, $l \in \{64, 32, 16\}$, on \texttt{cheetah\_run}, \texttt{quadruped\_run}, and \texttt{hopper\_hop}.
Director serves as the $\infty$-only baseline.
Table~\ref{tab:ablation_single} reports the final performance.
No single fixed resolution consistently matches MRS across tasks.
% Short-horizon skills ($l = 8, 16$) perform competitively on agility-demanding tasks but degrade on tasks requiring longer temporal commitment.
% Long-horizon skills ($l = 32, 64$) show the reverse pattern.
MRS, with access to the full resolution set, consistently outperforms all fixed alternatives, confirming that the gains arise from multi-resolution coverage as a structural inductive bias rather than from the presence of any single skill length.

\begin{table}[h]
    \centering
    \setlength{\tabcolsep}{5pt}
    \begin{tabular}{lccccccc}
        \toprule
        \textbf{Task} & \textbf{MRS} & $l=64$ & $l=32$ & $l=16$ & \textbf{$l=\infty$ (Director)} \\
        \midrule
        \texttt{cheetah\_run} & $866 \pm 29$ & $511 \pm 123$ & $329 \pm 41$ & $484 \pm 58$ & $435 \pm 129$ \\
        \texttt{quadruped\_run}  & $925 \pm 19$ & $729 \pm 70$ & $791 \pm 33$ & $701 \pm 63$ & $719 \pm 20$ \\
        \texttt{hopper\_hop} & $511 \pm 56$ & $195 \pm 37$ & $338 \pm 87$ & $378 \pm 83$ & $380 \pm 55$ \\
        \bottomrule
    \end{tabular}
    \caption{Single-resolution ablation. Final performance (mean $\pm$ std, $3$ seeds) for agents trained with a single fixed skill length versus full MRS. Director = $\infty$-only baseline.}
    \label{tab:ablation_single}
\end{table}

\vspace{-.5em}

\textbf{Meta-controller contribution.}
To isolate the contribution of the learned selection mechanism from the skill representations, we apply three alternative choice policies to a fully trained MRS agent at evaluation time: (a) the default learned meta-controller, (b) uniform random skill selection at each manager step, and (c) exclusive use of each individual skill head $\{64, 32, 16, 8, \infty\}$.
% All skills are trained on the same data that trains $N$ expert policy heads.
This setup compares the $N$ CVAEs and $N$ expert policy heads trained using the same data.
Table~\ref{tab:ablation_choice} reports mean scores across $100$ evaluation runs on all eight DMC tasks.
% We note that even when the CVAEs are trained on the same optimal data, skill interleaving with the meta-controller yields the best results.

\begin{table}[h]
    \centering
    \setlength{\tabcolsep}{3.5pt}
    \resizebox{\textwidth}{!}{%
    \begin{tabular}{lcccccccc}
        \toprule
        \textbf{Choice} & \texttt{walker\_walk} & \texttt{walker\_run} & \texttt{quad\_walk} & \texttt{quad\_run} & \texttt{cheetah\_run} & \texttt{hopper\_hop} & \texttt{cartpole\_sw} & \texttt{pendulum\_sw} \\
        \midrule
        Default & $\mathbf{967 \pm 20}$ & $\mathbf{771 \pm 12}$ & $\mathbf{942 \pm 29}$ & $\mathbf{900 \pm 38}$ & $\mathbf{861 \pm 15}$ & $\mathbf{497 \pm 61}$ & $\mathbf{825 \pm 113}$ & $\mathbf{526 \pm 424}$ \\
        Random & $961 \pm 20$ & $744 \pm 13$ & $881 \pm 39$ & $855 \pm 61$ & $729 \pm 86$ & $394 \pm 86$ & $517 \pm 269$ & $306 \pm 356$ \\
        $l=64$ & $965 \pm 21$ & $772 \pm 12$ & $929 \pm 28$ & $887 \pm 30$ & $784 \pm 40$ & $461 \pm 53$ & $678 \pm 253$ & $436 \pm 405$ \\
        $l=32$ & $962 \pm 20$ & $739 \pm 11$ & $937 \pm 30$ & $895 \pm 40$ & $687 \pm 86$ & $433 \pm 91$ & $614 \pm 302$ & $322 \pm 372$ \\
        $l=16$ & $958 \pm 23$ & $734 \pm 15$ & $787 \pm 52$ & $749 \pm 208$ & $483 \pm 35$ & $221 \pm 240$ & $474 \pm 272$ & $109 \pm 312$ \\
        $l=8$ & $953 \pm 30$ & $714 \pm 23$ & $698 \pm 64$ & $298 \pm 403$ & $544 \pm 100$ & $165 \pm 187$ & $231 \pm 89$ & $40 \pm 196$ \\
        $l=\infty$ & $952 \pm 25$ & $728 \pm 67$ & $879 \pm 37$ & $842 \pm 35$ & $846 \pm 17$ & $401 \pm 48$ & $573 \pm 248$ & $501 \pm 430$ \\
        \bottomrule
    \end{tabular}}
    \caption{Meta-controller ablation. Mean episodic reward across 100 evaluation runs. All conditions use the same trained MRS skill representations; only the choice mechanism varies. Best result per task in \textbf{bold}.}
    \label{tab:ablation_choice}
    \vspace{-.5em}
\end{table}

The default meta-controller achieves higher average performance than random selection and the best individual skill across tasks.
% The margin over the best individual skill is largest on tasks with the most varied dynamics, \texttt{quadruped\_run} and \texttt{hopper\_hop}, where switching between skill lengths within an episode is most consequential.
The margin over random selection confirms that the meta-controller has learned a structured, non-uniform selection policy.
Together, these results confirm that learned selection adds value beyond that provided by the individual skills.

\vspace{-.0em}

\begin{figure}[h]
    \centering
    \begin{subfigure}[b]{0.24\textwidth}
        \centering
        \includegraphics[width=\textwidth]{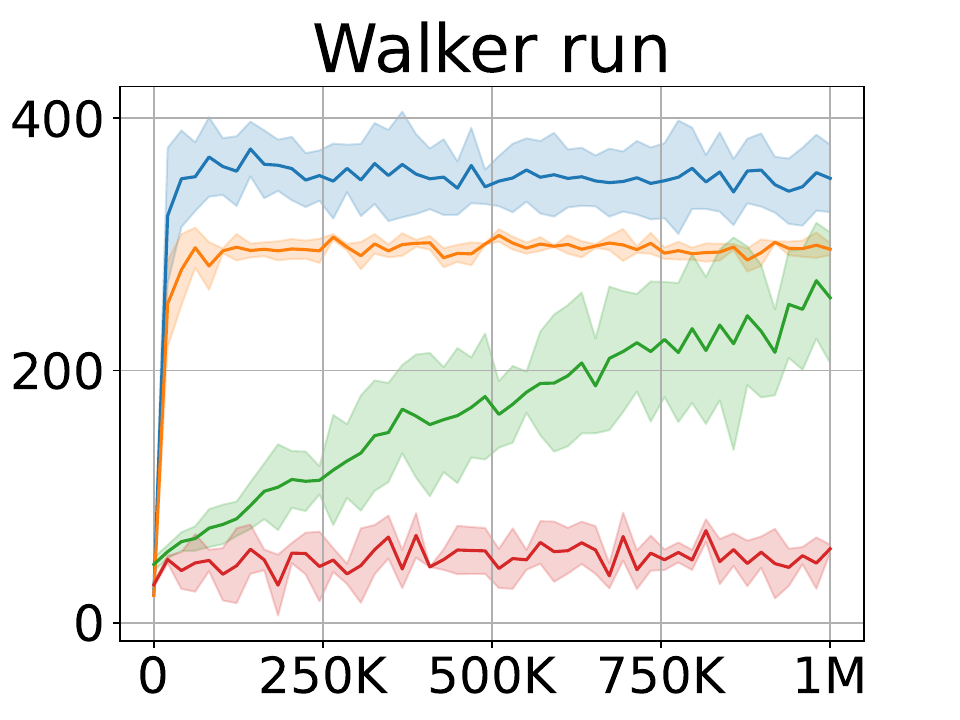}
    \end{subfigure}
    \begin{subfigure}[b]{0.24\textwidth}
        \centering
        \includegraphics[width=\textwidth]{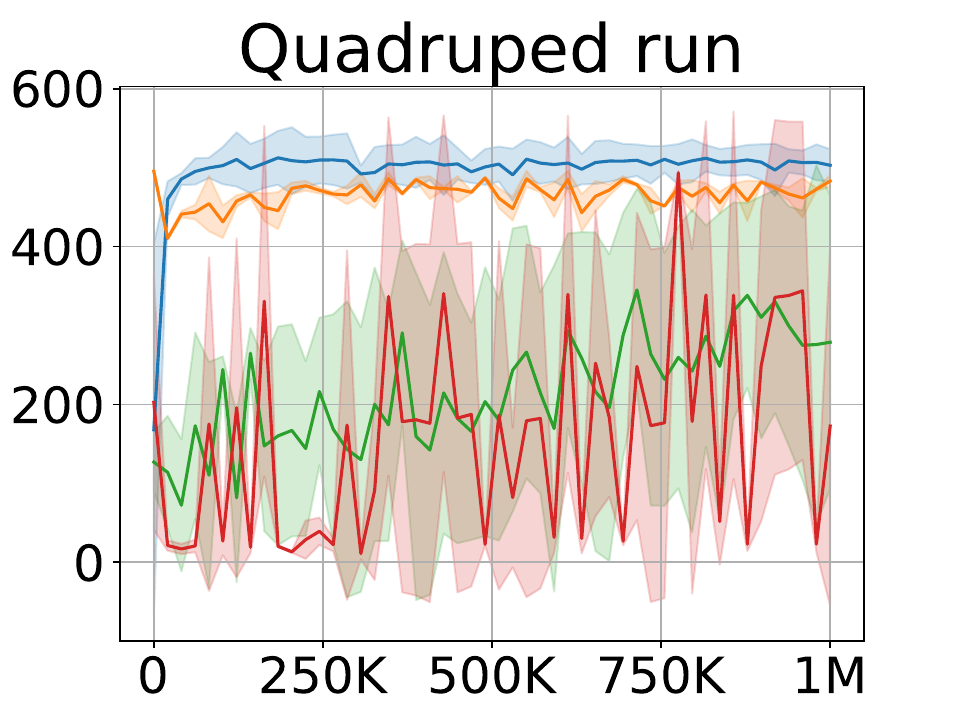}
    \end{subfigure}
    \begin{subfigure}[b]{0.24\textwidth}
        \centering
        \includegraphics[width=\textwidth]{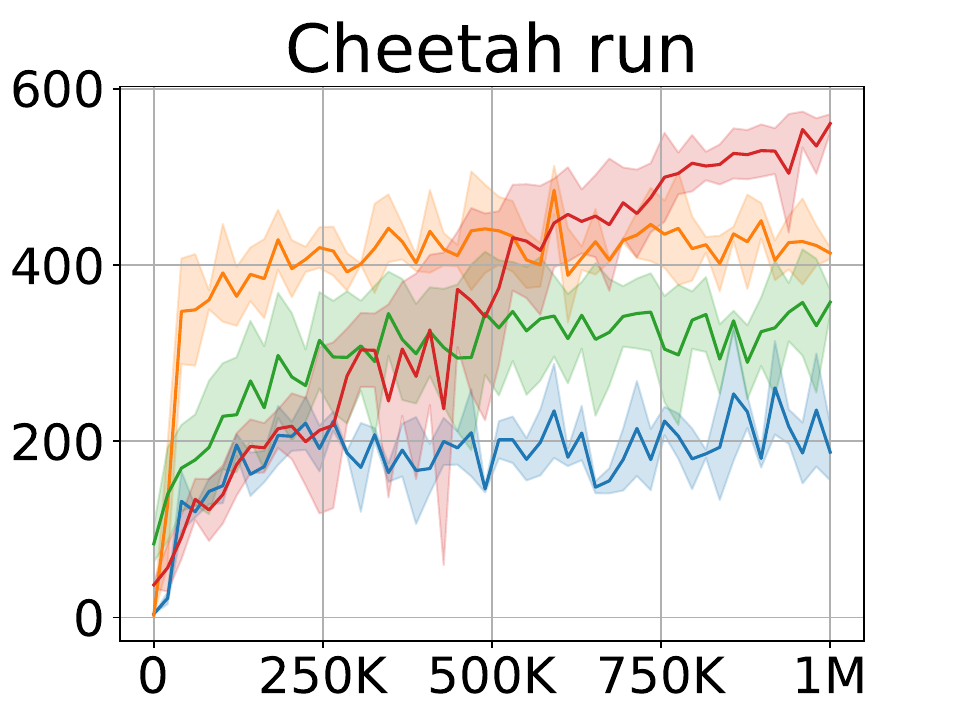}
    \end{subfigure}
    \begin{subfigure}[b]{0.24\textwidth}
        \centering
        \includegraphics[width=\textwidth]{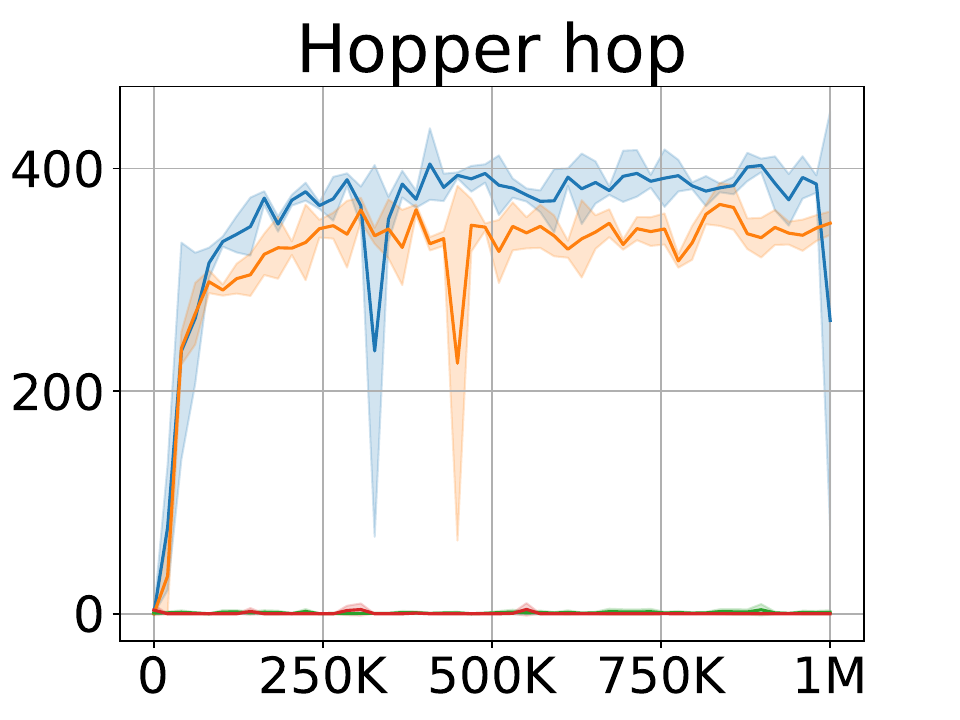}
    \end{subfigure}
    \begin{subfigure}[b]{0.4\textwidth}
        \centering
        \includegraphics[width=\textwidth]{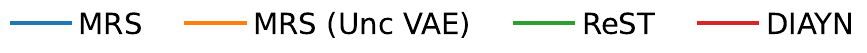}
    \end{subfigure}
    \caption{Performance comparison of agents fine-tuned for tasks after an exploration phase ($3$ seeds per experiment). The graphs show total episodic rewards (mean and standard deviation) against the global steps. The plots compare: MRS, MRS using exploratory rewards from the unconditional VAE, ReST, and DIAYN. Our agent is trained every $8$-th step using image inputs, while DIAYN/ReST trains every step using the internal environmental proprioceptive state.}
    \label{fig:finetune_scores}
    \vspace{-.5em}
\end{figure}

\textbf{Exploration objective.}
We compare four agents after an exploration phase followed by $1$M steps of task fine-tuning on four DMC tasks: MRS, MRS with a vanilla exploratory loss (replacing the $\min$-CVAE reconstruction error with Director's unconditional VAE reconstruction error), ReST \citep{jiang2022unsupervised}, and DIAYN \citep{eysenbach2018diversity}.
MRS operates on visual observations, whereas ReST and DIAYN use privileged ground-truth states.
MRS achieves competitive or superior fine-tuning performance and substantially faster convergence (Fig. \ref{fig:finetune_scores}).
% MRS with a vanilla exploratory loss is comparable to full MRS on most tasks but exhibits slower convergence, suggesting that the modified exploration objective contributes to diversity during the exploration phase, while the CVAE structure and temporal partitioning are the primary performance drivers.
MRS, with our exploratory objective, outperforms the vanilla training objective, indicating that the MRS exploratory objective alone can help learn useful state transitions. 

\vspace{-.5em}

\subsection{Qualitative Analysis}
\label{sec:qualitative}

\vspace{-.5em}

\begin{figure}[h]
    \centering
    \begin{subfigure}[b]{0.24\textwidth}
        \centering
        \includegraphics[width=\textwidth]{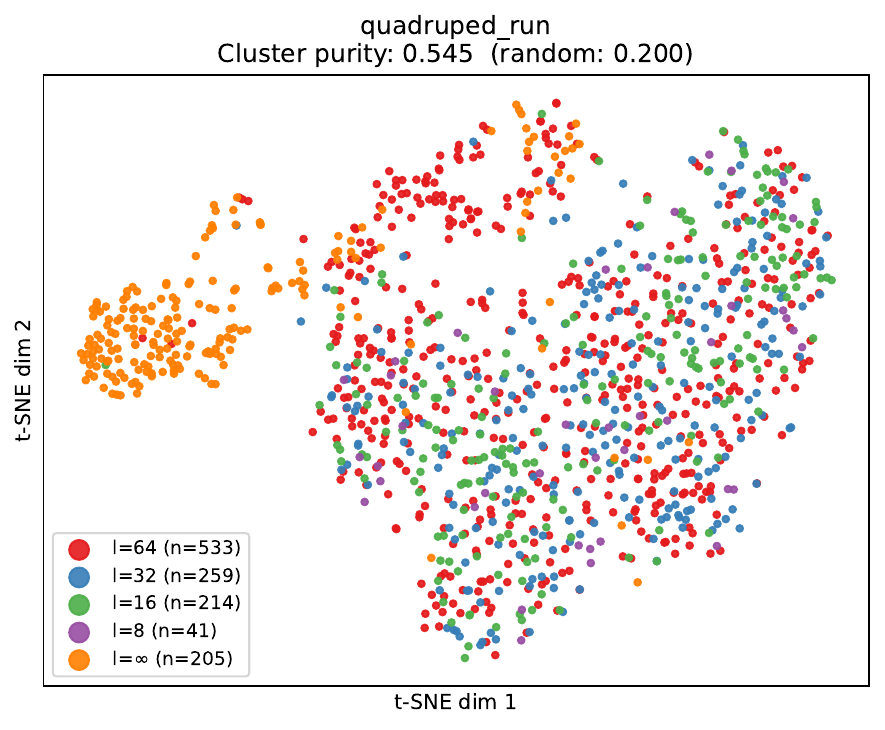}
    \end{subfigure}
    \begin{subfigure}[b]{0.24\textwidth}
        \centering
        \includegraphics[width=\textwidth]{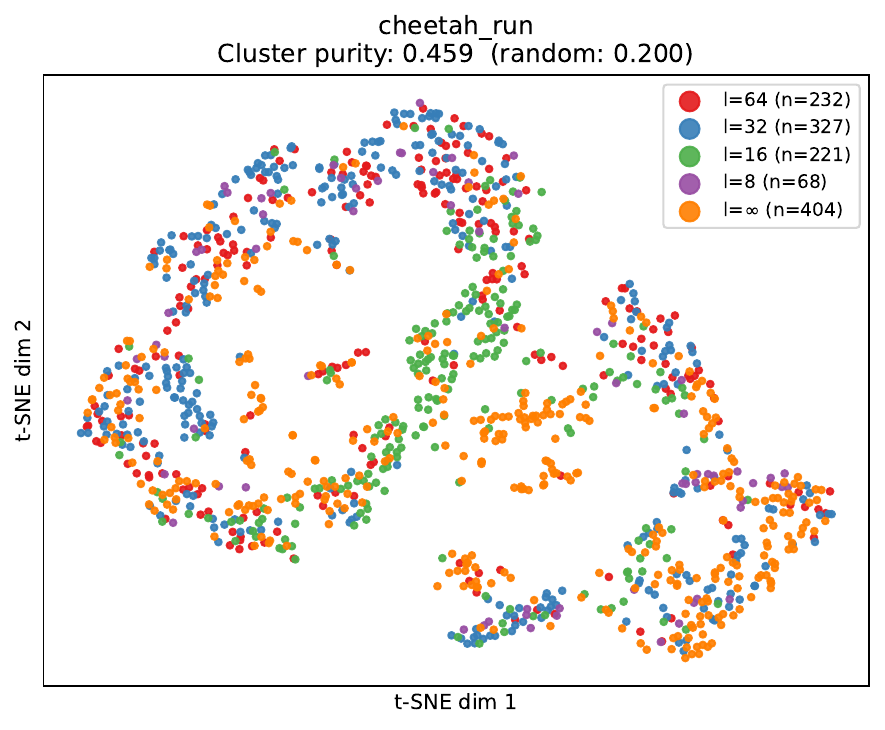}
    \end{subfigure}
    \begin{subfigure}[b]{0.24\textwidth}
        \centering
        \includegraphics[width=\textwidth]{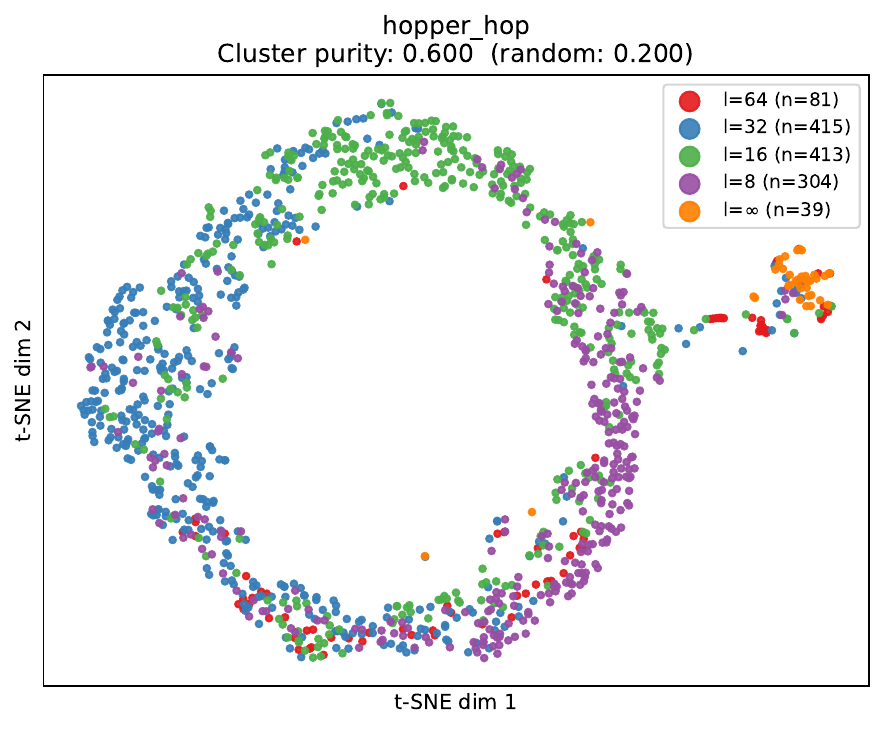}
    \end{subfigure}
    \begin{subfigure}[b]{0.24\textwidth}
        \centering
        \includegraphics[width=\textwidth]{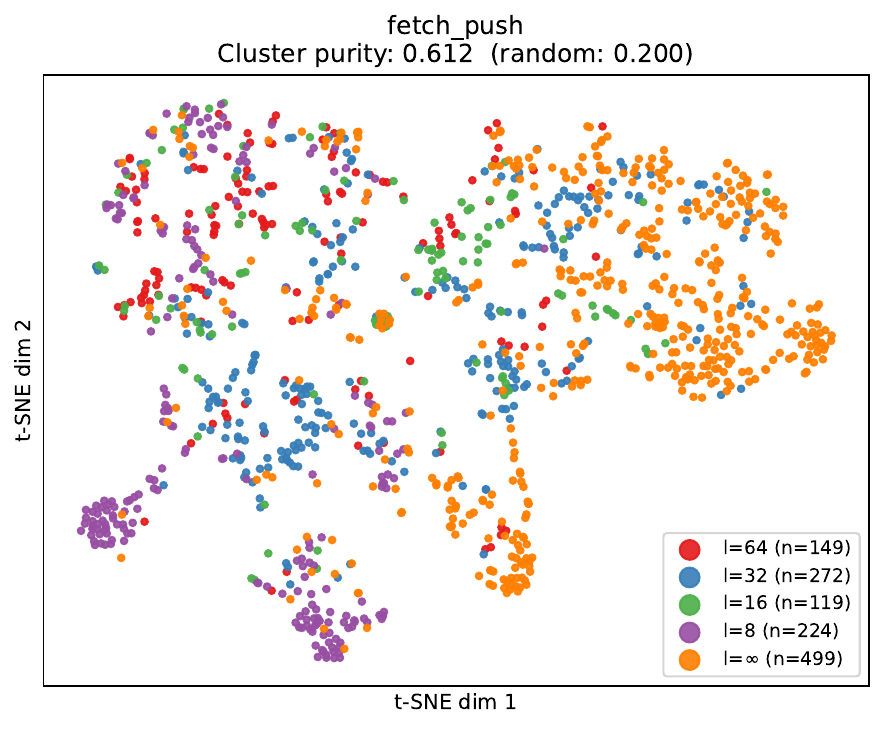}
    \end{subfigure}
    \caption{T-SNE projections of the environmental states, colored by the CVAE index $i$. The topological state clusters in the 2D space coincide with the color clusters, indicating that the agent prefers specific skill resolutions in specific state types. To measure cluster purity, we apply $k$-means clustering to the t-SNE projections, and purity is the ratio of the number of points assigned to the dominant skill to the total number of points in each cluster. Thus, following a pattern rather than being random.}
    \label{fig:choice_tsne}
    \vspace{-.5em}
\end{figure}

\textbf{Skill selection structure.}
Fig.~\ref{fig:choice_tsne} shows t-SNE projections of world model states visited during evaluation, colored by the skill length chosen by the meta-controller.
Across tasks, skill selection is spatially structured: distinct regions of the latent state space show consistent preferences for particular temporal horizons rather than uniform or random assignment across the embedding.
We quantify this with cluster purity: the fraction of points within each k-means cluster sharing the same dominant skill label, reporting a mean purity of $\mathbf{0.6}$ across tasks, substantially above the random baseline of $1/N = 0.20$.
This confirms that the meta-controller has learned a state-dependent selection policy with genuine structure, and supports the interpretation in Sec.~\ref{sec:main_results} that contextually appropriate subgoal selection in specific state regions drives the significant performance gains observed on tasks such as \texttt{cheetah\_run}.

\begin{figure}[h]
    \centering
    \begin{subfigure}[b]{0.24\textwidth}
        \centering
        \includegraphics[width=\textwidth]{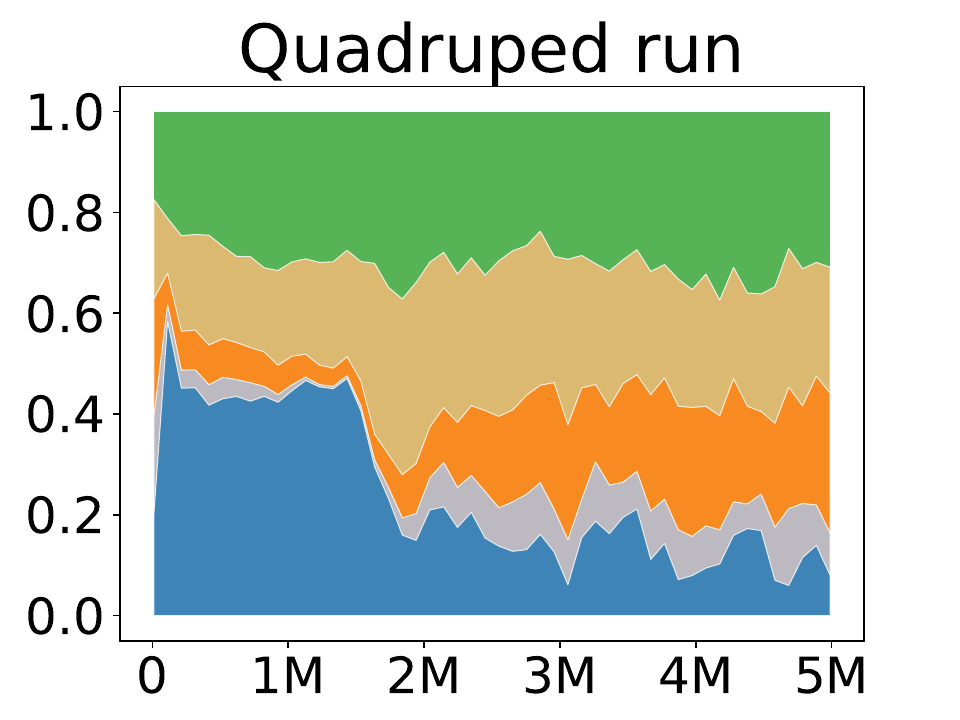}
    \end{subfigure}
    \begin{subfigure}[b]{0.24\textwidth}
        \centering
        \includegraphics[width=\textwidth]{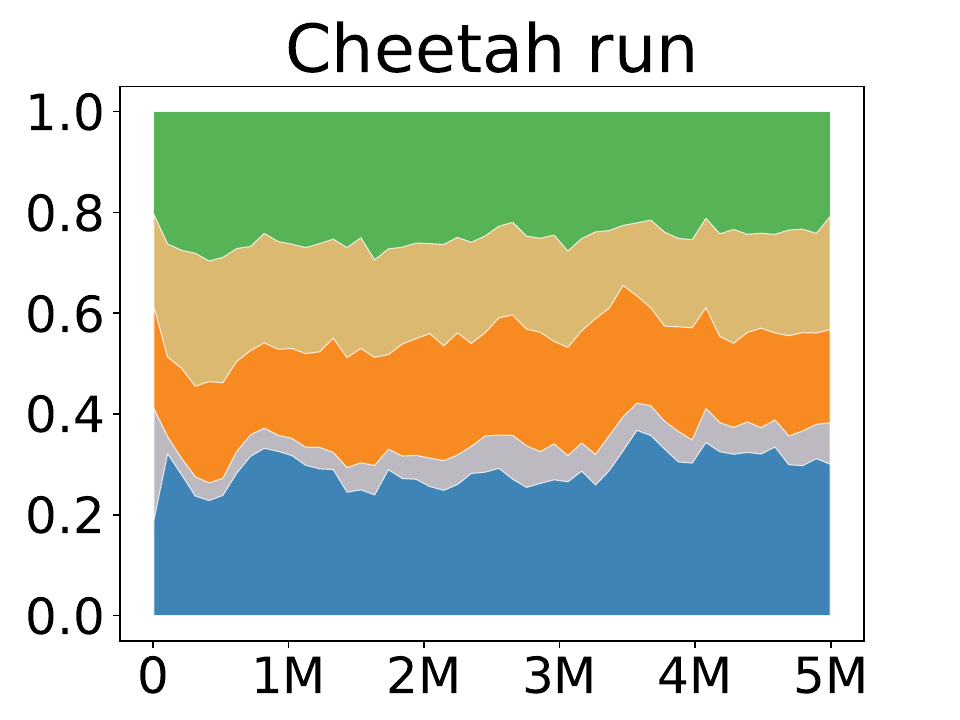}
    \end{subfigure}
    \begin{subfigure}[b]{0.24\textwidth}
        \centering
        \includegraphics[width=\textwidth]{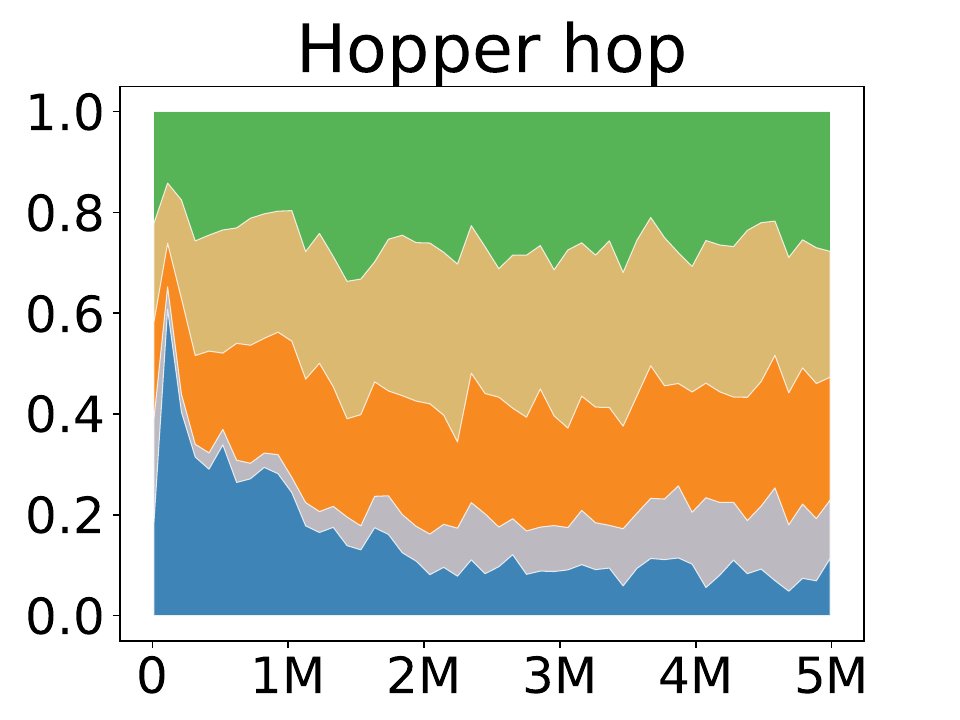}
    \end{subfigure}
    \begin{subfigure}[b]{0.24\textwidth}
        \centering
        \includegraphics[width=\textwidth]{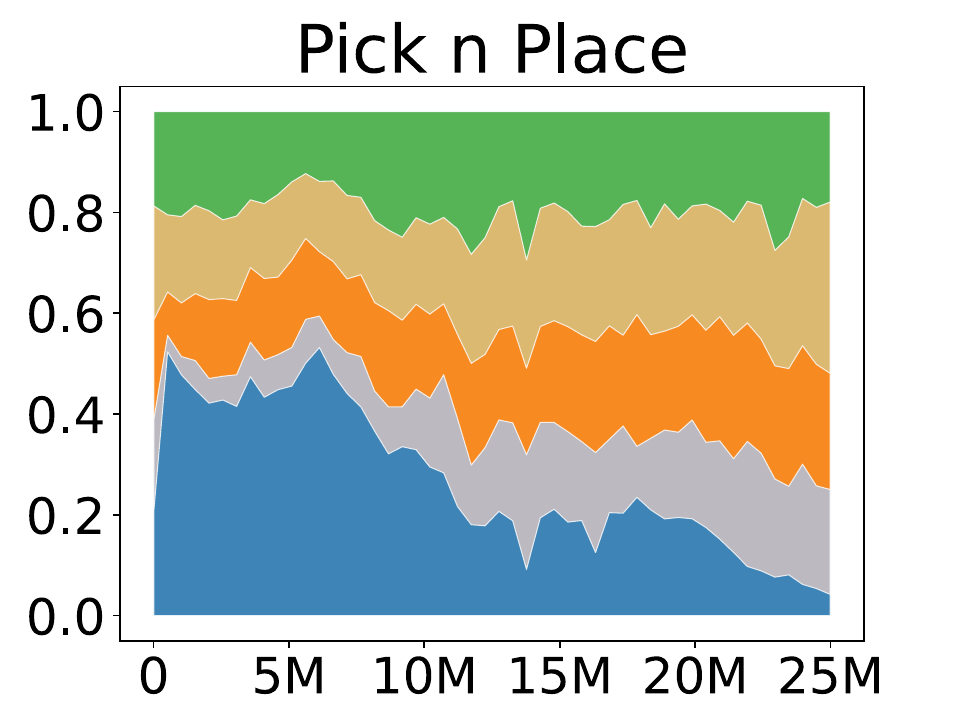}
    \end{subfigure}
    \begin{subfigure}[b]{0.5\textwidth}
        \centering
        \includegraphics[width=\textwidth]{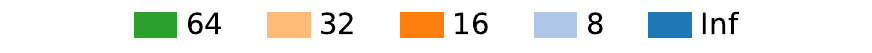}
    \end{subfigure}
    \caption{Stream graphs showing the evolution of the choice distribution during training averaged across $4$ seeds. A trend can be observed in some tasks where the manager initially focuses on $\infty$-length skills but gradually shifts to temporally constrained skills.}
    \label{fig:choice_hist}
    \vspace{-.5em}
\end{figure}

\textbf{Training dynamics of skill selection.}
Fig.~\ref{fig:choice_hist} shows the evolution of the choice distribution during training across all DMC and Gym-Robotics tasks.
A consistent pattern emerges: agents begin with a strong preference for the $\infty$-skill, then gradually shift toward finite skills as the CVAEs mature.
% This shift is most pronounced in tasks where precise subgoal selection is most valuable — hopper\_hop and quadruped\_run show the clearest transitions — and least pronounced in tasks where longer-horizon commitments remain useful throughout.
The initial $\infty$-preference is consistent with the chicken-and-egg motivation in Sec.~\ref{sec:mrsk}: early in training, finite CVAEs have not yet learned reliable transition representations, and the $\infty$-skill provides a stable source of exploratory subgoals.
As training proceeds and the CVAEs converge, the meta-controller progressively exploits the precision advantages of finite skills.

\textbf{Samples from trained models.}
We provide an extensive list of samples generated using a trained agent:
% Interesting behaviors like flips, maintaining headstand, and tumbling using the exploratory objective (Sec. \ref{sec:expl_skills}).
% Sample subgoal using each skill CVAE trained using task and exploratory rewards (Sec. \ref{sec:sample_goals}).
% Visual observations aggregate by choice (Sec. \ref{sec:states_per_choice}).

\begin{itemize}
    \item Interesting behaviors like flips, maintaining headstand, and tumbling using the exploratory objective (Sec. \ref{sec:expl_skills}).

    \item Sample subgoal using each skill CVAE trained using task and exploratory rewards (Sec. \ref{sec:sample_goals}).

    \item Visual observations aggregate by choice (Sec. \ref{sec:states_per_choice}).
\end{itemize}

\clearpage

%%%%%%%%%%%%%%%%%%%%%%%%%%%%%%%%%%%%%%%%%%%%%%%%%%%%%%%%%%%%%%%%
%% Appendices
%%%%%%%%%%%%%%%%%%%%%%%%%%%%%%%%%%%%%%%%%%%%%%%%%%%%%%%%%%%%%%%%
\appendix

\section{Appendix - Related Work}
\label{sec:related}

\vspace{-.5em}

\paragraph{Subgoal-based hierarchical RL.}
Hierarchical reinforcement learning formalizes temporal abstraction by allowing agents to select temporally-extended actions rather than only primitive actions \cite{sutton1999between, precup2000temporal, barto2003recent, botvinick2009hierarchically, pateria2021hierarchical}.
Early feudal decompositions \cite{dayan1992feudal} introduced the manager-worker structure that underlies modern HRL; neural instantiations include FeUdal Networks \cite{vezhnevets2017feudal} and h-DQN \cite{kulkarni2016hierarchical}, which demonstrated that learned subgoal hierarchies could tackle tasks with sparse rewards.
The Option-Critic architecture \cite{bacon2017option} extended this line by learning option policies and termination conditions end-to-end without manually specified subtasks.
Goal-conditioned methods such as HIRO \cite{nachum2018data} and Director \cite{hafner2022deep} train a higher-level policy to propose subgoal states and a lower-level policy to reach them, enabling long-horizon planning from pixels without access to privileged state information.
A parallel line of work focuses on improving the quality of subgoal representations directly: HRAC \cite{zhang2020generating} constrains subgoals to an adjacency-based reachable set, and HLPS \cite{wang2024probabilistic} introduces probabilistic subgoal representations to improve robustness under stochasticity.
These methods address the unconstrained goal space in isolation but do not partition the goal representation by temporal horizon, and therefore cannot resolve the precision-smoothness tradeoff that MRS identifies as a structural source of the HRL agility gap.

\vspace{-.5em}

\paragraph{Temporal abstraction and variable commitment.}
A distinct line of work addresses how long a higher-level commitment should be maintained.
Fixed-commitment methods \cite{florensa2017stochastic, gupta2018meta} hold a skill active for a predetermined number of steps, sacrificing adaptability for simplicity.
Timed subgoals \cite{gurtler2021hierarchical} associate each subgoal with a fixed commitment window, improving temporal structure but providing no mechanism to vary commitment based on local state context.
HiPPO \cite{li2019sub} explicitly tackles fixed temporal abstraction by randomizing the commitment length at each manager step, sampled from a categorical distribution over predetermined periods; this improves robustness to environment changes by averaging over the nuisance of skill duration rather than learning to exploit it.
STRAW \cite{vezhnevets2016strategic} learns macro-actions of variable length by predicting when to replan, but does so reactively rather than by maintaining a structured set of temporal resolutions.
More recent work introduces reachability-aware abstractions \cite{zadem2024reconciling} that refine the goal space iteratively based on $k$-step reachability relations, providing spatial structure but without an explicit temporal partitioning mechanism.
MRS takes a different stance from all of these: rather than fixing, randomizing, or reactively terminating temporal commitments, MRS maintains a discrete set of skills at geometrically spaced temporal horizons and learns a meta-controller that selects among them based on current state, treating temporal resolution as a structured, learnable signal rather than a nuisance variable.

\vspace{-.5em}

\paragraph{Unsupervised skill discovery.}
A large body of work discovers diverse skills without external rewards by maximizing mutual information (MI) between a latent skill variable and states or trajectories.
Representative methods include DIAYN \cite{eysenbach2018diversity}, which maximizes MI between skill codes and visited states; DADS \cite{sharma2020dynamics}, which maximizes MI between skills and state transition dynamics; VALOR \cite{achiam2018variational} and OPAL \cite{ajay2020opal}, which extend MI objectives to full trajectory segments; and InfoGAN-based approaches \cite{kurutach2018learning}. Sequential coverage methods such as ReST \cite{jiang2022unsupervised} build on this by training skills sequentially to increase state-space coverage.
The principal advantage of these methods is behavioral diversity without reward engineering; their principal limitation is that mutual-information objectives impose no temporal structure on the skill space and provide no mechanism for state-conditioned prediction of reachable future states at a specific horizon.
Because they do not condition on $s_t$ or constrain skill length, they cannot concentrate representational capacity on locally reachable transitions as MRS's CVAEs do.
MRS lies between unsupervised skill discovery and goal-conditioned HRL: it retains the goal-conditioned framework to enable temporal structure, while its exploratory objective draws on the diversity-seeking motivation of skill-discovery methods.
This connection is evaluated directly in Sec.~\ref{sec:ablations}, where MRS's exploration phase is compared against DIAYN and ReST after fine-tuning on downstream tasks.

\section{Appendix - Discussion}
\label{sec:discussion}

\vspace{-.5em}

The results establish that the HRL agility gap is not primarily a capacity or training problem, because MRS uses comparable parameters to Director and is trained at the same frequency, yet consistently reduces the gap with DreamerV3 on agility-demanding tasks while retaining competitiveness on long-horizon and sparse-reward benchmarks.
The ablations support a more specific conclusion: the gap is a representational problem arising from the absence of a structure in the goal space.
Imposing a structure, like temporal, through fixed-horizon CVAEs and a learned discrete gate, without modifying the world model, worker, or training procedure, is sufficient to produce the observed gains.
This has implications beyond MRS: it suggests that other subgoal-based HRL architectures may benefit from similar temporal partitioning of their goal or option spaces, independent of the specific world model or policy optimization algorithm used.
Two limitations qualify these conclusions.
First, MRS inherits Director's dependence on a reliable RSSM.
Tasks where world model quality is low, the CVAEs will also struggle, and the precision advantage of constrained subgoal prediction may be reduced.
Second, the resolution set $\{K, 2K, 4K, 8K, \infty\}$ is fixed across all tasks; while the geometric spacing is principled, the optimal set may be task-dependent, and MRS currently has no mechanism to discover or adapt it.

Three directions follow naturally from these observations.
First, adaptive resolution discovery learning the resolution set rather than fixing it, for instance by initializing with a dense set of horizons and pruning underused skills during training, would remove the remaining hand-specified structure from MRS and potentially improve performance on tasks where the geometric spacing is suboptimal.
Second, the architecture naturally accommodates hybrid skills that combine learned CVAEs with analytic or hand-specified expert primitives, which may be valuable in robotics settings where certain motion primitives are known a priori and do not need to be learned from data.
Third, the multi-head policy gradient formulation is architecture-agnostic: the discrete choice gate and per-head gradient separation could be applied to other HRL frameworks beyond Director, including options-based hierarchies \cite{bacon2017option} and language-conditioned managers \cite{jiang2019language}, extending the benefits of temporal partitioning to a broader class of methods.

\clearpage

%%%%%%%%%%%%%%%%%%%%%%%%%%%%%%%%%%%%%%%%%%%%%%%%%%%%%%%%%%%%%%%%
%% NOTE: THIS MARKS THE END OF THE "MAIN TEXT"
%%%%%%%%%%%%%%%%%%%%%%%%%%%%%%%%%%%%%%%%%%%%%%%%%%%%%%%%%%%%%%%%

%%%%%%%%%%%%%%%%%%%%%%%%%%%%%%%%%%%%%%%%%%%%%%%%%%%%%%%%%%%%%%%%
%% Bibliography
%%%%%%%%%%%%%%%%%%%%%%%%%%%%%%%%%%%%%%%%%%%%%%%%%%%%%%%%%%%%%%%%
\clearpage
\bibliography{main}
\bibliographystyle{rlj}

%%%%%%%%%%%%%%%%%%%%%%%%%%%%%%%%%%%%%%%%%%%%%%%%%%%%%%%%%%%%%%%%
% AUTHOR: If your paper has no supplementary materials, you may 
%         comment out the line below, which creates the title for
%         the supplementary materials.
%%%%%%%%%%%%%%%%%%%%%%%%%%%%%%%%%%%%%%%%%%%%%%%%%%%%%%%%%%%%%%%%
\beginSupplementaryMaterials

% Content that appears after the references are not part of the ``main text,'' have no page limits, are not necessarily reviewed, and should not contain any claims or material central to the paper. 
% %
% If your paper includes supplementary materials, use the \begin{center}
%     {\tt {\textbackslash}beginSupplementaryMaterials} 
% \end{center}
% command as in this example, which produces the title and disclaimer above. 
% %
% If your paper does not include supplementary materials, this command can be removed or commented out.

\section{Derivation of Policy Gradients for MRS}
\label{app:mrs_policy_grads}

We first decompose the action prediction process to derive the policy gradient to train the manager and worker policies.
Let an MRS agent be in state $s_t$ at step $t$.
Every $K$-th step, the manager refreshes the worker's goal.
For clarity, let the abstract step be indexed by $k$, then at each abstract step ($t=kK$):

\begin{enumerate}
    \item Sample skill latents from the skill heads: $z_{k,0},z_{k,1},...,z_{k,N-1} \sim \Pi^{N-1}_{i=0} \pi_{M_i}(z_{k,i}|s_{kK})$.
    \item Sample a choice variable: $c_k \sim \pi_{M_C}(c_k|s_{kK})$.
    \item Compute the selected subgoal: $s^k_g = \sum_{i=0}^{N-1} c_{k,i} \cdot \text{Dec}^i_\phi(s_{kK},z_{k,i})$.
    \item Predict the environmental actions using worker: $\pi_W(a_t|s_t,s_g^k)$
\end{enumerate}

Thus, the trajectory probability that starts at $s_0$ can be written as:

\begin{equation}
    p(\tau) = p(s_0) \prod_{k=0}^{\floor{T/K}-1}
    \underbrace{\pi_{M_C}(c_k|s_{kK}) \prod_{i=0}^{N-1} \pi_{M_i}(z_{k,i}|s_{kK})^{c_{k,i}}}_{\text{Manager}}
    \prod_{t=0}^{T-1}    
    \underbrace{\pi_W(a_t|s_t,s_g^{\floor{t/K}})}_{\text{Worker}}
    \cdot
    \underbrace{p_T(s_{t+1}|a_t,s_t)}_{\text{State transition}}
\end{equation}

The components of the equation can be read as: the manager predicts the skills $(z_{k,0},z_{k,1},...,z_{k,N-1})$ and choice $c_k$ for every abstract step $k$, the worker predicts the action $a_t$ at each step $t$ using the subgoal $s_g^{\floor{t/K}}$ for the duration, and the environmental state transition $p_T$.
Here, the exponent $c_{k,i}$ collapses the probability $\pi_{M_i}(z_{k,i}|s_{kK})$ of the unselected skill at step $kK$ head to $1$, since it does not affect the trajectory.

We follow the policy gradient derivation from \cite{sutton2018reinforcement}.
The aim is to compute $\nabla_\theta J$, where $J = \mathbb{E}_\tau[R(\tau)]$ is the expected reward and $\theta$ are the policy parameters.
Using the standard log-derivative trick (\cite{sutton2018reinforcement}), the objective can be written as maximizing the trajectory log-probability weighted by the expected reward:

\begin{gather*}
    \nabla_\theta J = \mathbb{E}_\tau [R(\tau) \cdot \nabla_\theta\log p(\tau)]
\end{gather*}

The gradient of the trajectory log-probability w.r.t. the manager parameters $M$ is:

\begin{align*}
    \nabla_M \log p(\tau) &= \sum_{k=0}^{\floor{T/K}-1} [\nabla_M\log \pi_{M_C}(c_{k}|s_{kK}) + \sum_{i=0}^{N-1}c_{k,i} \nabla_M \log \pi_{M_i}(z_{k,i}|s_{kK})]
\end{align*}

Therefore, the policy-gradient objective can be written as:

\begin{gather*}
    \nabla_M J = \mathbf{E}_\tau [R(\tau) \cdot \sum_{k=0}^{\floor{T/K}-1} [\nabla_M\log \pi_{M_C}(c_{k}|s_{kK}) + \sum_{i=0}^{N-1}c_{k,i} \nabla_M \log \pi_{M_i}(z_{k,i}|s_{kK})]]
\end{gather*}

Given these policy gradients, we construct the losses for each head as the sum of the policy gradient objective and an entropy maximization objective (Eq. \ref{eq:mrsk_skill_loss},\ref{eq:mrsk_choice_loss}), and sum them for the total loss (Eq. \ref{eq:mrsk_policy_loss}).

\begin{gather}
    \label{eq:lambda_returns}
    G_k^\lambda = R_k + \gamma ((1 - \lambda)v_M(s_{kK}) + \lambda G^\lambda_{k+1}) \\
    \label{eq:mrsk_choice_loss}
    \mathcal{L}(\pi_{M_c}) = -\mathbb{E}_\tau \sum_{k=0}^{\floor{T/K}-1} \log \pi_{M_c}(c_k|s_{kK}) \cdot A_k + \eta \mathbb{H}[\pi_{M_C}(c_k|s_{kK})] \\
    \label{eq:mrsk_skill_loss}
    \mathcal{L}(\pi_{M_i}) = -\mathbb{E}_\tau \sum_{k=0}^{\floor{T/K}-1} c_{k,i} \cdot \log \pi_{M_i}(z_{k,i}|s_{kK}) \cdot A_k + \eta \mathbb{H}[\pi_{M_i}(z_{k,i}|s_{kK})] \\
    \label{eq:mrsk_policy_loss}
    \mathcal{L}(\pi_M) = \mathcal{L}(\pi_{M_c}) + \sum_{i=0}^{N-1} \mathcal{L}(\pi_{M_i}) \\
    \label{eq:mrsk_critic_loss}
    \mathcal{L}(v_M) = \mathbb{E}_\tau \sum_{k=0}^{\floor{T/K}-1} ( v_M(s_{kK}) - G^\lambda_k )^2
\end{gather}

Where $A_k = G^\lambda_k - v_M(s_{kK})$ is the lambda returns estimated using abstract trajectories (Eq. \ref{eq:lambda_returns}), $v_M$ is the critic (Eq. \ref{eq:mrsk_critic_loss}).
The policy maximizes the advantage $G^\lambda_k - v_M(s_{kK})$ instead of directly maximizing estimated returns.
Weighted entropic losses $\mathbb{H}[\cdot]$ encourage adequate exploration prior to convergence.
The manager estimates separate returns, learns separate critics, and computes separate advantages for the external rewards $R^{Extr}_k$ and exploratory rewards $R^{Expl}_k$ (Eq. \ref{eq:expl_rew}).
The rewards are summed over the $K$-length trajectory chunks.
The total advantage is the weighted sum of the exploratory and external advantages ($([1.0, 0.1])$ in our case).

\clearpage

\section{Full Results}
\label{app:full_train_curves}

\subsection{Training Curves}
\label{sec:train_curves}

\begin{figure}[h]
    \centering
    \begin{subfigure}[b]{0.24\textwidth}
        \centering
        \includegraphics[width=\textwidth]{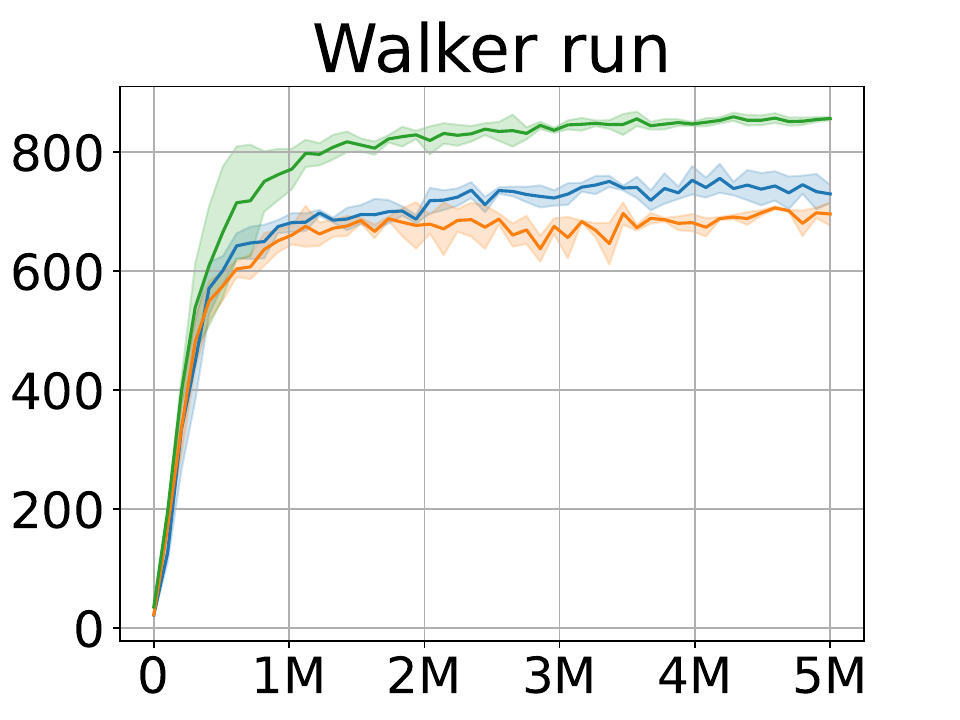}
    \end{subfigure}
    \begin{subfigure}[b]{0.24\textwidth}
        \centering
        \includegraphics[width=\textwidth]{figures/plots/main_scores/dmc_quadruped_run.pdf}
    \end{subfigure}
    \begin{subfigure}[b]{0.24\textwidth}
        \centering
        \includegraphics[width=\textwidth]{figures/plots/main_scores/dmc_cheetah_run.pdf}
    \end{subfigure}
    \begin{subfigure}[b]{0.24\textwidth}
        \centering
        \includegraphics[width=\textwidth]{figures/plots/main_scores/dmc_hopper_hop.pdf}
    \end{subfigure}
    \begin{subfigure}[b]{0.24\textwidth}
        \centering
        \includegraphics[width=\textwidth]{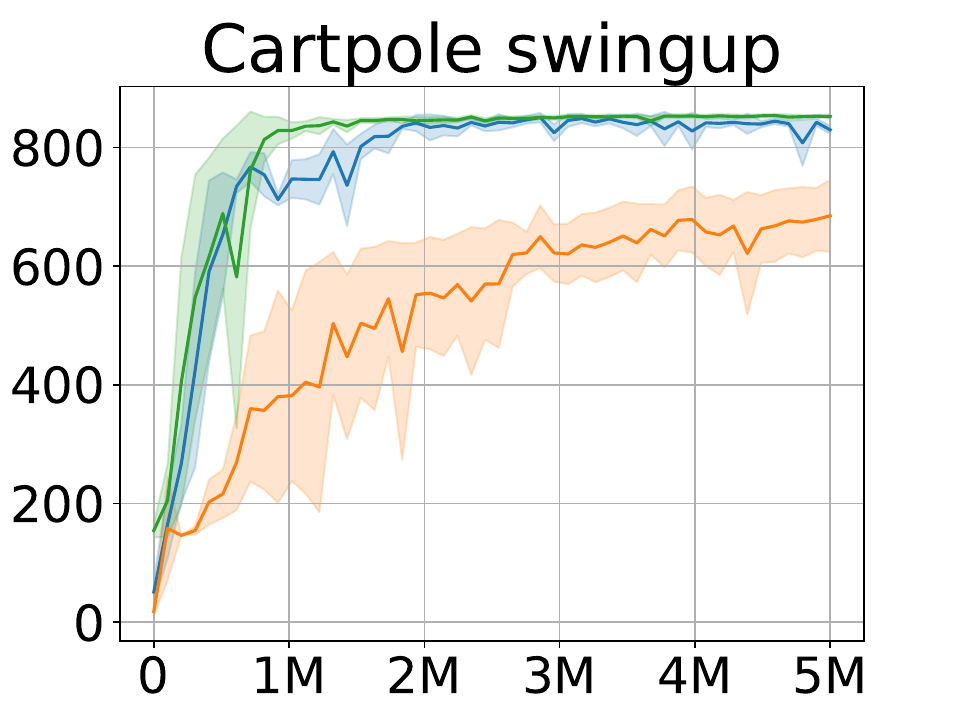}
    \end{subfigure}
    \begin{subfigure}[b]{0.24\textwidth}
        \centering
        \includegraphics[width=\textwidth]{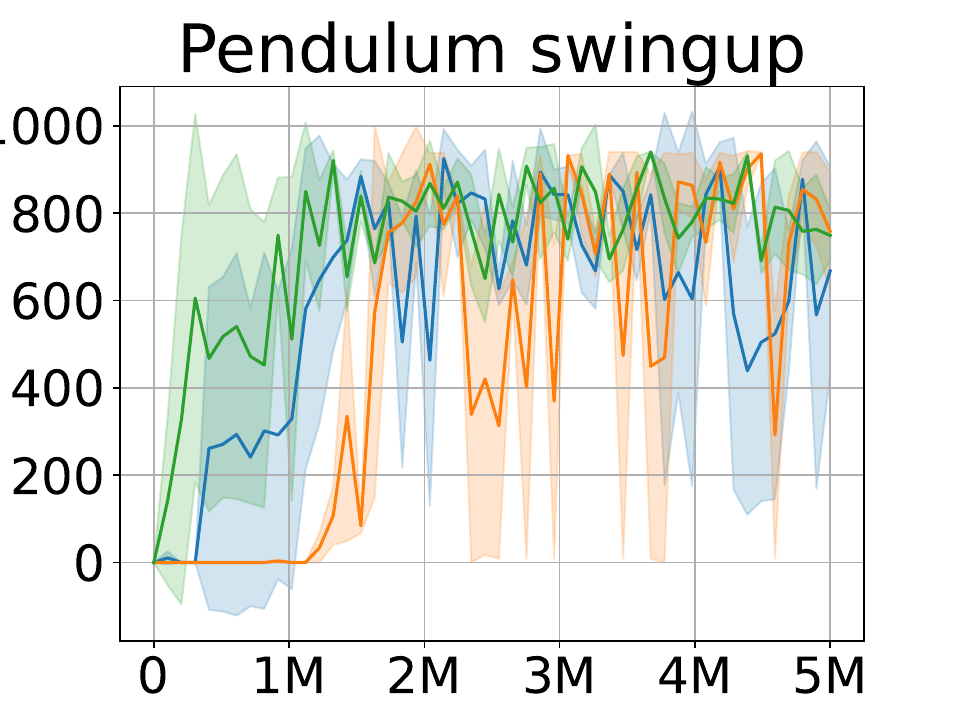}
    \end{subfigure}
    \begin{subfigure}[b]{0.24\textwidth}
        \centering
        \includegraphics[width=\textwidth]{figures/plots/main_scores/gymrobo_fetch_push.pdf}
    \end{subfigure}
    \begin{subfigure}[b]{0.24\textwidth}
        \centering
        \includegraphics[width=\textwidth]{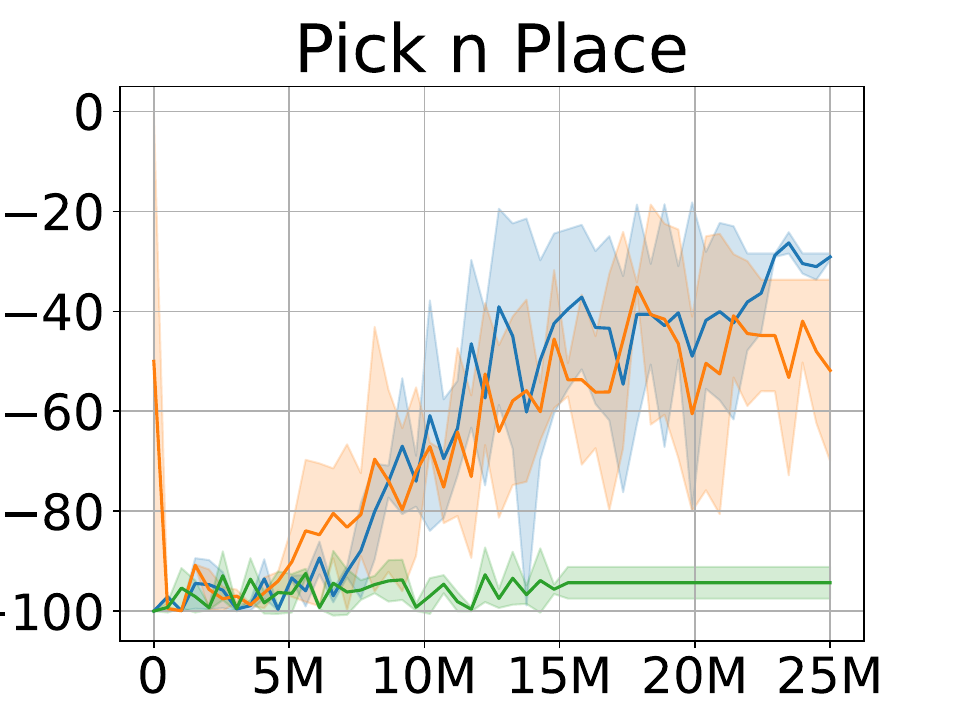}
    \end{subfigure}
    \begin{subfigure}[b]{0.48\textwidth}
        \centering
        \includegraphics[width=\textwidth]{figures/plots/main_scores/dmc_scores_legend.pdf}
    \end{subfigure}
    \caption{Episode scores from MRS (ours), Director, and DreamerV3 ($4$ seeds per experiment). The plot shows the total rewards (mean and standard deviation) received in an episode against the environmental step. Both the Director and MRS use the same common hyperparameters. The first \textit{six} tasks are from the DMC suite, and the last \textit{two} (Push, and Pick n Place) are from the Gymnasium-Robotics suite. It can be seen that MRS noticeably improves the base model's performance in all cases while maintaining computational efficiency.}
    \label{fig:app_main_results}
\end{figure}

\subsection{t-SNE Projections}

\begin{figure}[h]
    \centering
    \begin{subfigure}[b]{0.24\textwidth}
        \centering
        \includegraphics[width=\textwidth]{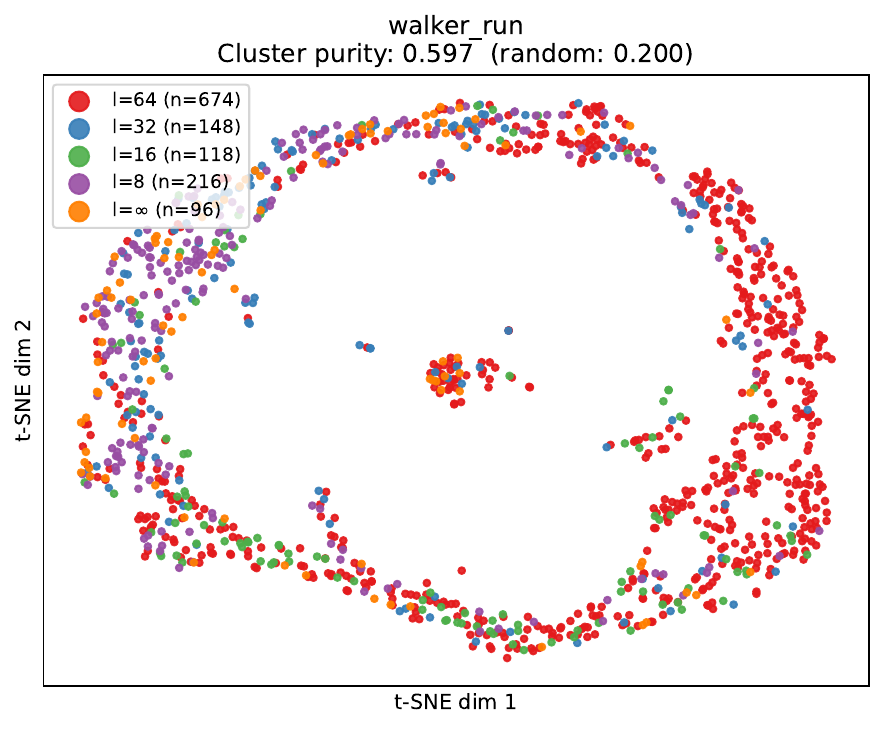}
    \end{subfigure}
    \begin{subfigure}[b]{0.24\textwidth}
        \centering
        \includegraphics[width=\textwidth]{figures/plots/choice_tsne/dmc_quadruped_run_tsne.pdf}
    \end{subfigure}
    \begin{subfigure}[b]{0.24\textwidth}
        \centering
        \includegraphics[width=\textwidth]{figures/plots/choice_tsne/dmc_cheetah_run_tsne.pdf}
    \end{subfigure}
    \begin{subfigure}[b]{0.24\textwidth}
        \centering
        \includegraphics[width=\textwidth]{figures/plots/choice_tsne/dmc_hopper_hop_tsne.pdf}
    \end{subfigure}
    \begin{subfigure}[b]{0.24\textwidth}
        \centering
        \includegraphics[width=\textwidth]{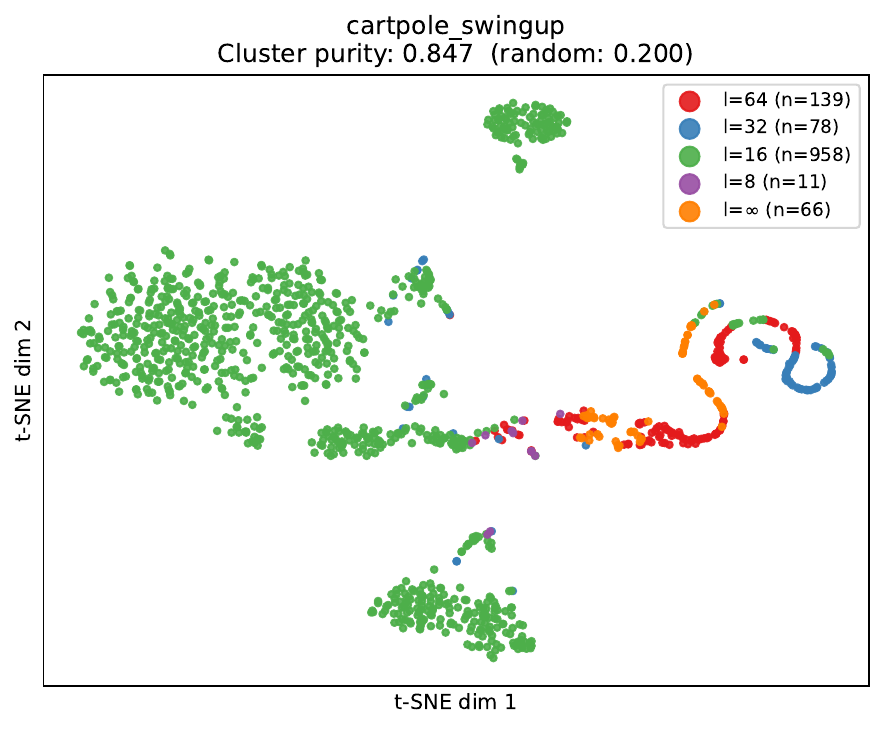}
    \end{subfigure}
    \begin{subfigure}[b]{0.24\textwidth}
        \centering
        \includegraphics[width=\textwidth]{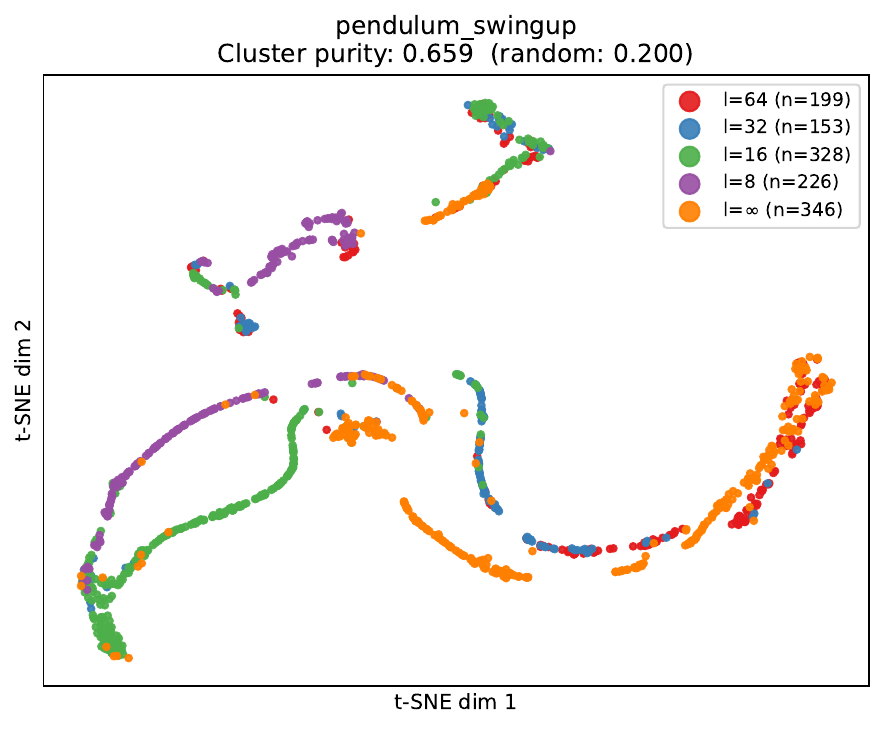}
    \end{subfigure}
    \begin{subfigure}[b]{0.24\textwidth}
        \centering
        \includegraphics[width=\textwidth]{figures/plots/choice_tsne/gymrobo_fetch_push_tsne.pdf}
    \end{subfigure}
    \begin{subfigure}[b]{0.24\textwidth}
        \centering
        \includegraphics[width=\textwidth]{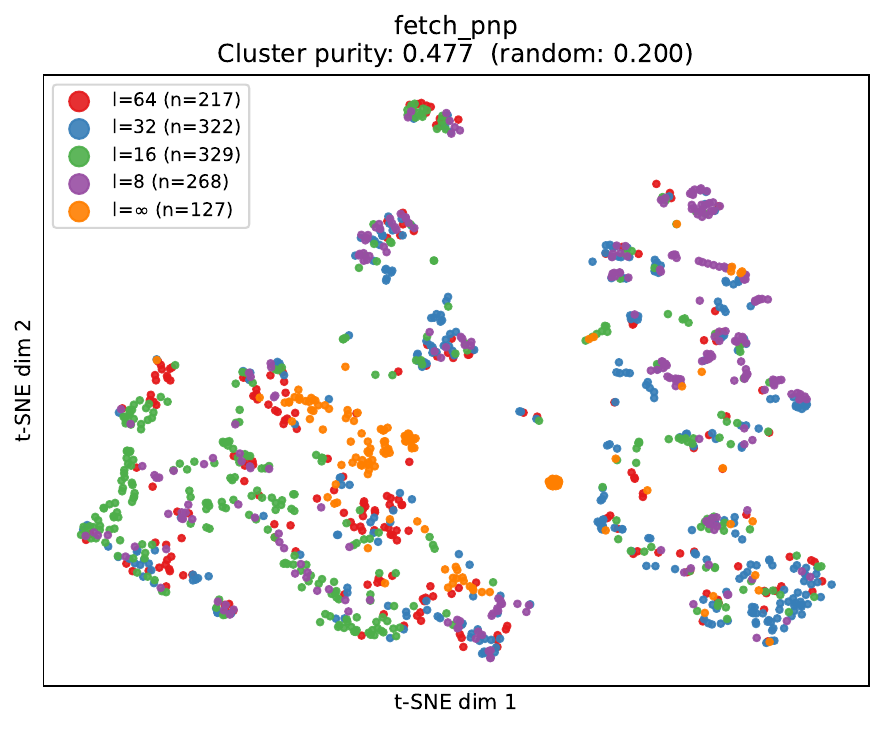}
    \end{subfigure}
    \caption{T-SNE projections of the states, colored by the chosen skill's index for the DMC and Gym-Robotics tasks. To measure cluster purity, we apply $k$-means clustering (with the number of clusters chosen to maximize the silhouette score), and the purity is the ratio of the number of points belonging to the dominant skill to the total points in each cluster. Thus, following a pattern instead of being random.}
    \label{fig:app_choice_tsne}
    % \vspace{-1.em}
\end{figure}

\subsection{Choice Streamgraphs}

\begin{figure}[h]
    \centering
    \begin{subfigure}[b]{0.24\textwidth}
        \centering
        \includegraphics[width=\textwidth]{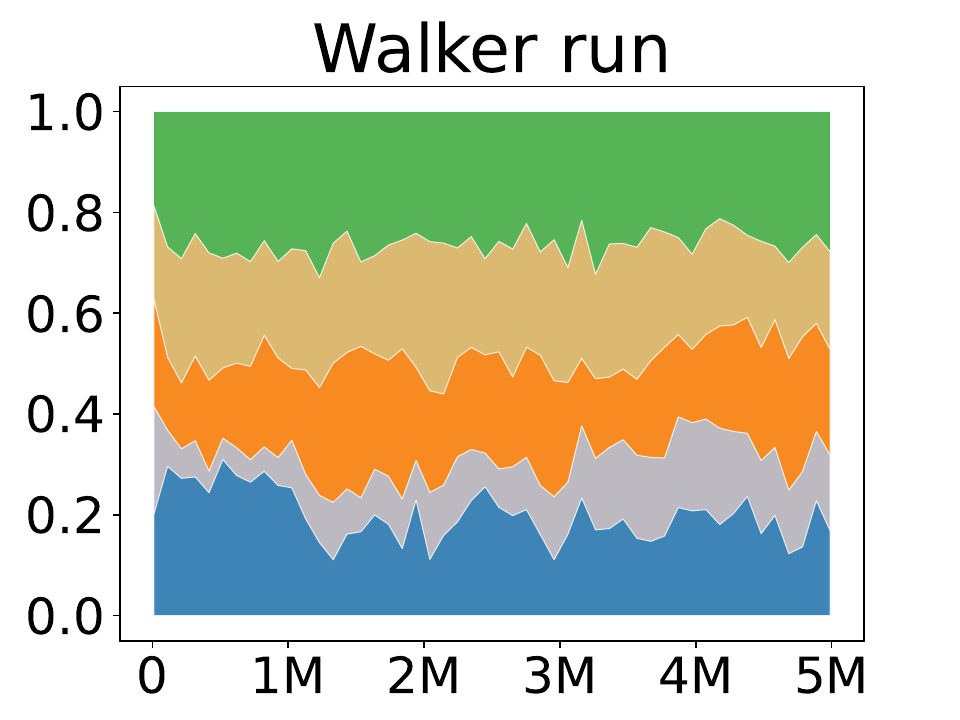}
    \end{subfigure}
    \begin{subfigure}[b]{0.24\textwidth}
        \centering
        \includegraphics[width=\textwidth]{figures/plots/choice_hist/dmc_quadruped_run_choice_hist.pdf}
    \end{subfigure}
    \begin{subfigure}[b]{0.24\textwidth}
        \centering
        \includegraphics[width=\textwidth]{figures/plots/choice_hist/dmc_cheetah_run_choice_hist.pdf}
    \end{subfigure}
    \begin{subfigure}[b]{0.24\textwidth}
        \centering
        \includegraphics[width=\textwidth]{figures/plots/choice_hist/dmc_hopper_hop_choice_hist.pdf}
    \end{subfigure}
    \begin{subfigure}[b]{0.24\textwidth}
        \centering
        \includegraphics[width=\textwidth]{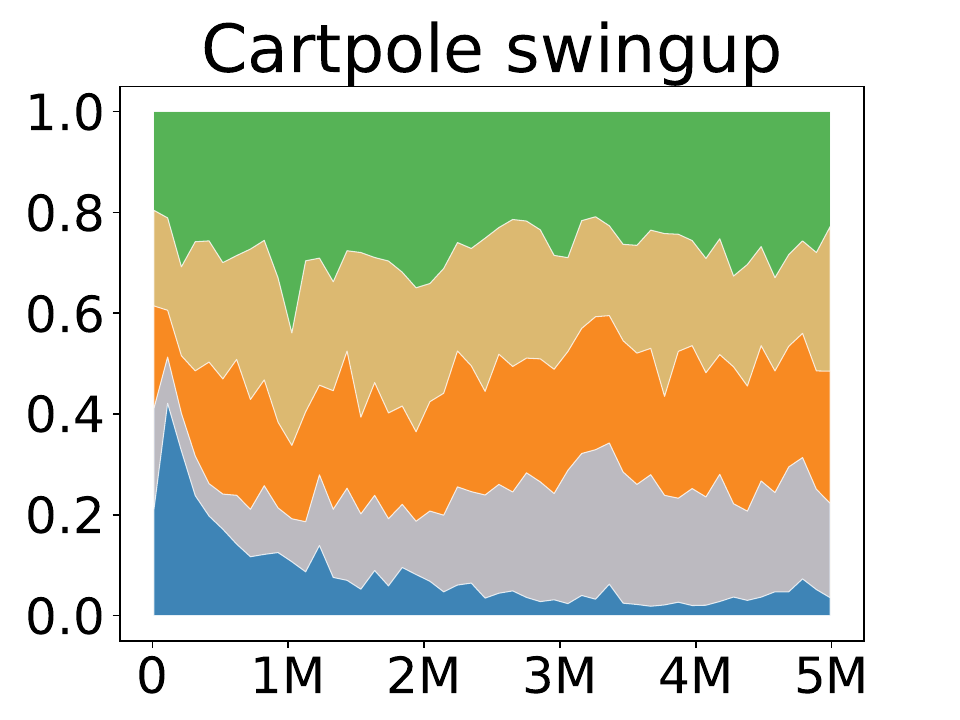}
    \end{subfigure}
    \begin{subfigure}[b]{0.24\textwidth}
        \centering
        \includegraphics[width=\textwidth]{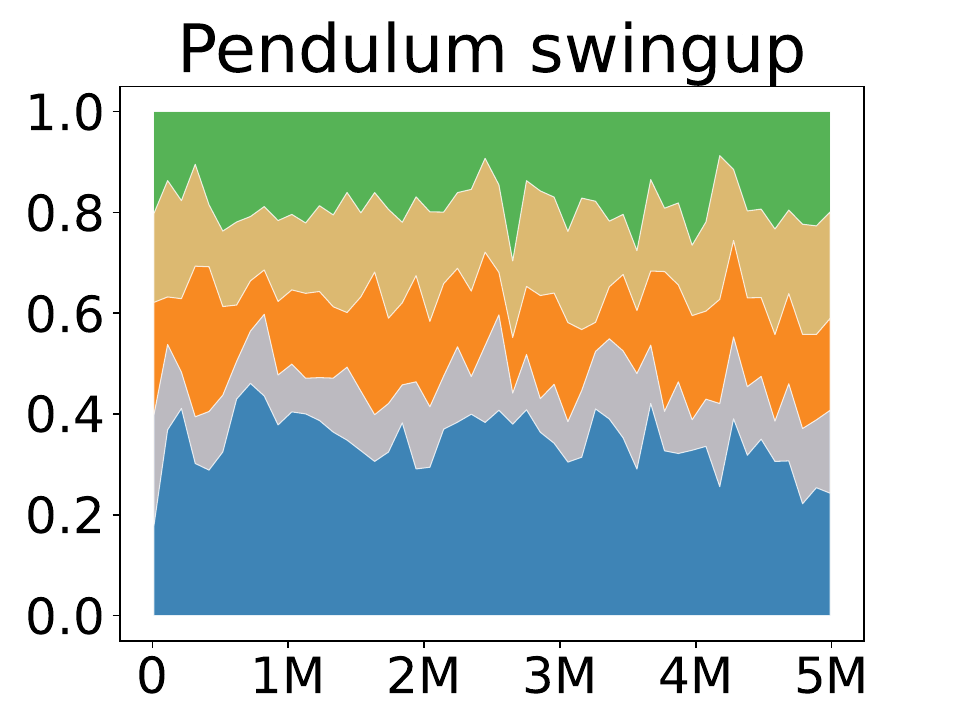}
    \end{subfigure}
    \begin{subfigure}[b]{0.24\textwidth}
        \centering
        \includegraphics[width=\textwidth]{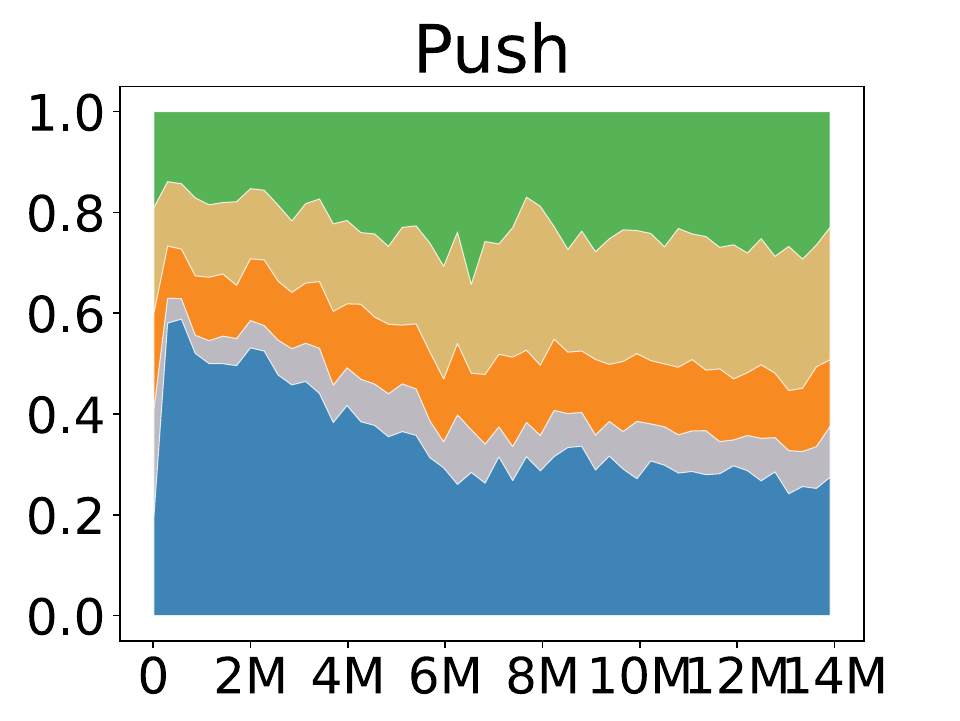}
    \end{subfigure}
    \begin{subfigure}[b]{0.24\textwidth}
        \centering
        \includegraphics[width=\textwidth]{figures/plots/choice_hist/gymrobo_fetch_pnp_choice_hist.pdf}
    \end{subfigure}
    \begin{subfigure}[b]{0.5\textwidth}
        \centering
        \includegraphics[width=\textwidth]{figures/plots/choice_hist/choice_hist_legend.pdf}
    \end{subfigure}
    \caption{Stream graphs showing the evolution of the choice distribution during training averaged across $4$ seeds. A trend can be observed in some tasks where the manager initially focuses on $\infty$-length skills but gradually shifts to temporally constrained skills.}
    \label{fig:app_choice_hist}
    % \vspace{-1.em}
\end{figure}

\clearpage

\section{Architecture \& Training Details}
\label{sec:app_training_details}

\subsection{Worker}
\label{sec:worker_training}

The worker is trained using $K$-step imagined rollouts $(\kappa \sim \pi_W)$.
% Given rewards $R_i$, computed using the CSR modules for the explorer, or as the cosine-similarity measures for the worker.
Given the imagined trajectory $\kappa$, the rewards for the worker $R^W_t$ are computed as the $\texttt{cosine\_max}$ similarity measure (Eq. \ref{eq:cosine_max}) between the trajectory states $s_t$ and the prescribed worker goal $s_g$ (Eq. \ref{eq:worker_rew}).
First, discounted returns $G^\lambda_t$ are computed as $n$-step lambda returns (Eq. \ref{eq:worker_lambda_return}).
Then the Actor policy is trained using the SAC objective (Eq. \ref{eq:worker_actor_loss}) and the Critic is trained to predict the discounted returns (Eq. \ref{eq:worker_critic_loss}).
The entropy for the worker and the manager is weighted to maintain a target entropy.

\begin{gather}
    \label{eq:cosine_max}
    \texttt{cosine\_max}(a,b)=||a \cdot b||/\max(||a||,||b||)^2 \\
    \label{eq:worker_rew}
    R^W_{t} = \texttt{cosine\_max}(s_t,s_g) \\
    \label{eq:worker_lambda_return}
    G^\lambda_t = R^W_{t+1} + \gamma_L ((1 - \lambda)v(s_{t+1}) + \lambda G^\lambda_{t+1}) \\
    \label{eq:worker_actor_loss}
    \mathcal{L}(\pi_W) = -\mathbb{E}_{\kappa \sim \pi_W} \sum_{t=0}^{H-1} \left[ (G^\lambda_t - v_W(s_t)) \ln \pi_W(z|s_t) + \eta \text{H}[\pi_W(z|s_t)] \right] \\
    \label{eq:worker_critic_loss}
    \mathcal{L}(v_W) = \mathbb{E}_{\kappa \sim \pi_W} \left[ \sum_{t=0}^{H-1} (v_W(s_t) - G^\lambda_t)^2 \right]
\end{gather}

\subsection{Implementation Details}

We implement two functions: \texttt{policy} (Alg. \ref{alg:mrsk_policy}) and \texttt{train} (Alg. \ref{alg:mrsk_train}), using the hyperparameters shown in Table \ref{tab:hyperparam}.
The functions are implemented in Python/Tensorflow using XLA JIT compilation.
The experiments on average take $2$ days to run $5$M steps on an NVIDIA RTX 5000.

\begin{table}[h]
    \centering
    \begin{tabular}{|c|c|c|}
        \hline\hline
        Name & Symbol & Value \\
        \hline\hline
        Train batch size & $B$ & $16$ \\
        Replay data length & - & $64$ \\
        Worker abstraction length & $K$ & $8$ \\
        Explorer Imagination Horizon & $T$ & $16$ \\
        Return Lambda & $\lambda$ & $0.95$ \\
        Return Discount & $\gamma$ & $0.99$ \\
        Skill resolutions & $L$ & $\{64,32,16,8,\infty\}$ \\
        Target entropy & $\eta$ & 0.5 \\
        KL loss weight & $\beta$ & 1.0 \\
        RSSM deter size & - & $1024$ \\
        RSSM stoch size & - & $32 \times 32$ \\
        Optimizer & - & Adam \\
        Learning rate (all) & - & $10^{-4}$ \\
        Adam Epsilon & - & $10^{-6}$ \\
        Weight decay (all) & - & $10^{-2}$ \\
        Activations & - & LayerNorm + ELU \\
        MLP sizes & - & $4 \times512$ \\
        Train every & - & 8 \\
        Prallel Envs & - & 4 \\
        \hline
    \end{tabular}
    \caption{Agent Hyperparameters}
    \label{tab:hyperparam}
\end{table}

\begin{algorithm}[H]
    \DontPrintSemicolon
    \KwIn{Collected trajectories $\mathcal{D} = \{\tau_1,...,\tau_B\}$}
    % \KwOut{Updated world model $\texttt{wm}$, skill modules $(\text{Enc}_\phi, \text{Dec}_\phi)$, manager $\pi_M$, worker $\pi_W$}
    
    \BlankLine
    \tcp{World Model Training}
    % $\theta_W \leftarrow \argmin_\theta \mathbb{E}_{\tau \sim \mathcal{D}}[\mathcal{L}_{WM}(\tau;\theta)]$ 
    $\text{\texttt{wm}.train}(\mathcal{D})$ \tcp*{See \cite{hafner2019learning}}
    
    \BlankLine
    \tcp{Multi-Resolution Skill Learning}
    $\mathcal{L}_{\text{skills}} \leftarrow [\;]$\;
    \For{$l_i \in \mathcal{L}$}{
        $\{(s_t, s_{t+l_i})\} \leftarrow \text{ExtractStatePairs}(\mathcal{D}, l_i)$ 
        
        $\mathcal{L}_i \leftarrow \text{skill\_loss}(s_t,s_{t+l_i}) $\tcp*{CVAE loss (Eq. \ref{eq:skvae_loss})}
        $\mathcal{L}_{\text{skills}}.\text{append}(\mathcal{L}_i)$\;
    }
    $\text{update\_skills}(\text{sum}(\mathcal{L}_{\text{skills}}))$\;
    
    \BlankLine
    \tcp{Policy Optimization via Imagination}
    $\mathcal{S}_{init} \leftarrow \{s_0\ |\ s_0 \in \tau, \tau \in \mathcal{D}\}$ \tcp*{Initial states}
    $\hat{\tau} \leftarrow \texttt{wm.imagine}(\pi_\text{MRS}, \mathcal{S}_{init}, T)$ \tcp*{Rollout imagined trajectories}
    
    \BlankLine
    \tcp{Reward Computation}
    $\hat{\tau}.r^{extr} \leftarrow r_{env}(\hat{\tau})$ \tcp*{Environment reward}
    $\hat{\tau}.r^{expl} \leftarrow \text{expl\_rew}(\hat{\tau})$ \tcp*{Exploration reward (Eq. \ref{eq:expl_rew})}
    $\hat{\tau}.r^{goal} \leftarrow \texttt{cosine\_max}(\hat{\tau}.s_t, \hat{\tau}.s^{\floor{t/K}}_g)$ \tcp*{Goal achievement reward}
    
    \BlankLine
    \tcp{Hierarchical Policy Update}
    $\mathcal{T}_W \leftarrow \texttt{split}(\hat{\tau})$ \tcp*{Worker-level transitions}
    $\mathcal{T}_M \leftarrow \texttt{abstract}(\hat{\tau})$ \tcp*{Manager-level abstractions}

    $\mathcal{L}(\pi_M), \mathcal{L}(v_M) = \text{manager\_loss}(\mathcal{T}_M)$ \tcp*{Eqs. \ref{eq:mrsk_policy_loss},\ref{eq:mrsk_critic_loss}}
    $\text{update\_manager}(\mathcal{L}(\pi_M), \mathcal{L}(v_M))$

    $\mathcal{L}(\pi_W), \mathcal{L}(v_W) = \text{worker\_loss}(\mathcal{T}_W)$ \tcp*{Eqs. \ref{eq:worker_actor_loss},\ref{eq:worker_critic_loss}}
    $\text{update\_worker}(\mathcal{L}(\pi_W), \mathcal{L}(v_W))$

    % $\theta_W \leftarrow \theta_W - \nabla_{\theta} \mathcal{L}_{actor}(\mathcal{T}_W;\theta)$ \tcp*{Eq. \ref{eq:worker_loss}}
    % $\theta_M \leftarrow \theta_M - \nabla_{\theta} \mathcal{L}_{manager}(\mathcal{T}_M;\theta)$ \tcp*{Eq. \ref{eq:manager_loss}}
    
    \caption{Multi-Resolution Skill Training}
    \label{alg:mrsk_train}
\end{algorithm}

\begin{algorithm}[H]
    \DontPrintSemicolon
    \KwIn{Observation $o_t$, Agent previous state $(t, s_{t-1}, a_{t-1}, s_g)$}
    \KwOut{Action $a_t$, New agent state $(t+1, s_t, a_t, s_g)$}
    
    \BlankLine
    % \textbf{State Representation:} \\
    $s_t \leftarrow \texttt{wm}(o_t, s_{t-1}, a_{t-1})$ \tcp*{World model state computation}
    
    \BlankLine
    \If{$t \bmod K = 0$}{
        \tcp{Manager updates goal every $K$ steps}
        $(z_0,z_1,...,z_{N-1}, c) \sim {\pi_M}(s_t)$ \tcp*{Sample skill latent $z$ and choice $c$}
        
        % \tcp{Multi-resolution skill decoding}
        $\{s_g^i\}_{i=0}^{N-1} \leftarrow \{\text{Dec}^i_\phi(s_t,z_i)\}_{i=0}^{N-1}$ \tcp*{Generate candidate goals}
        
        $s_g \leftarrow \sum_{i=0}^{N-1} c_i \cdot s_g^i$ \tcp*{Select goal using choice vector $c$}
    }
    \Else{
        $s_g \leftarrow s_g$ \tcp*{Persist previous goal}
    }
    
    \BlankLine
    \tcp{Worker policy execution}
    $a_t \leftarrow \text{Worker}_\pi(s_t, s_g)$ \tcp*{Generate action for current goal}
    
    \BlankLine
    % \textbf{State Transition:} \\
    Return $a_t, (t+1, s_t, a_t, s_g)$
    
    \caption{Multi-Resolution Skill Policy ($\pi_\text{MRS}$)}
    \label{alg:mrsk_policy}
\end{algorithm}

\clearpage

\section{Behaviors Learned via Exploration}
\label{sec:expl_skills}

\begin{figure}[h]
    \centering
    \begin{subfigure}[b]{\textwidth}
        \centering
        \includegraphics[width=\textwidth]{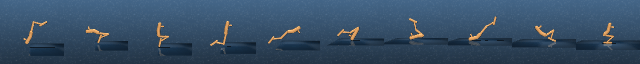}
        % \caption{Hopper learns to use a front flip to stand.}
    \end{subfigure}
    \begin{subfigure}[b]{\textwidth}
        \centering
        \includegraphics[width=\textwidth]{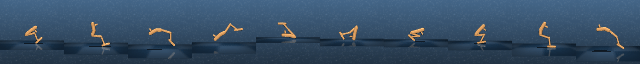}
        \caption{Hopper learns to use a front flip to stand, and back flips.}
    \end{subfigure}
    \begin{subfigure}[b]{\textwidth}
        \centering
        \includegraphics[width=\textwidth]{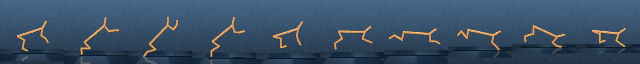}
        % \caption{Cheetah learns to leap forward.}
    \end{subfigure}
    \begin{subfigure}[b]{\textwidth}
        \centering
        \includegraphics[width=\textwidth]{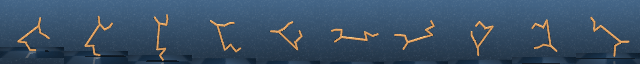}
        \caption{Cheetah learns to leap forward and perform perfect back flips.}
    \end{subfigure}
    \begin{subfigure}[b]{\textwidth}
        \centering
        \includegraphics[width=\textwidth]{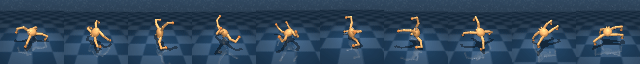}
        % \caption{Side rolls learned by the Quadruped.}
    \end{subfigure}
    \begin{subfigure}[b]{\textwidth}
        \centering
        \includegraphics[width=\textwidth]{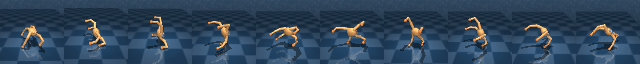}
        \caption{Quadruped learning side rolls and walking on two legs.}
    \end{subfigure}
    \begin{subfigure}[b]{\textwidth}
        \centering
        \includegraphics[width=\textwidth]{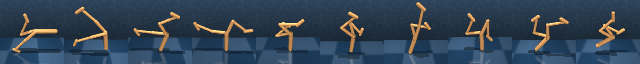}
        % \caption{Side rolls learned by the Quadruped.}
    \end{subfigure}
    \begin{subfigure}[b]{\textwidth}
        \centering
        \includegraphics[width=\textwidth]{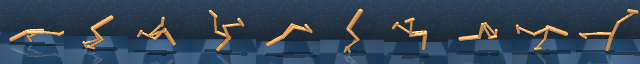}
        \caption{Walker trying to maintain a headstand repeatedly and fast-forward tumbling using head and legs.}
    \end{subfigure}
    \caption{Samples of some movements learned and regularly performed by the agent optimized only for the exploratory loss.}
    \label{fig:expl_skill_samples}
\end{figure}

We observed several interesting behaviors exhibited by the MRS agent during training solely with the exploratory loss, including front flips, back flips, and jumps.
The intrinsic exploratory loss encourages the agent to perform novel state transitions (Sec. \ref{sec:training}).
Fig. \ref{fig:expl_skill_samples} shows some of the learned movements.

\clearpage

\section{Sample Goals using Skill CVAEs}
\label{sec:sample_goals}

We generate some sample goals using each of the Skill CVAEs individually.
The goals are generated using a uniform prior $p(z)$ for the skills, and initial states $s_t$ sampled from the replay database.
We use skills with temporal resolutions $[64,32,16,8,\infty]$, but we omit the $\infty$ length skills, as they correspond to simply learning all states independently and are not our contribution.

\subsection{Default Objective}

\begin{figure}[h]
    \centering
    \begin{subfigure}[b]{\textwidth}
        \centering
        \includegraphics[width=\textwidth]{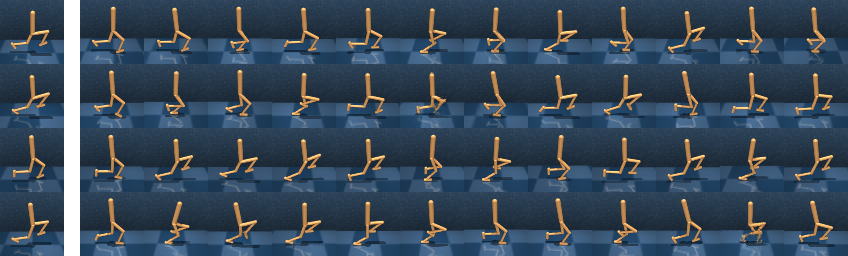}
        \caption{$64$ length.}
    \end{subfigure}
    \begin{subfigure}[b]{\textwidth}
        \centering
        \includegraphics[width=\textwidth]{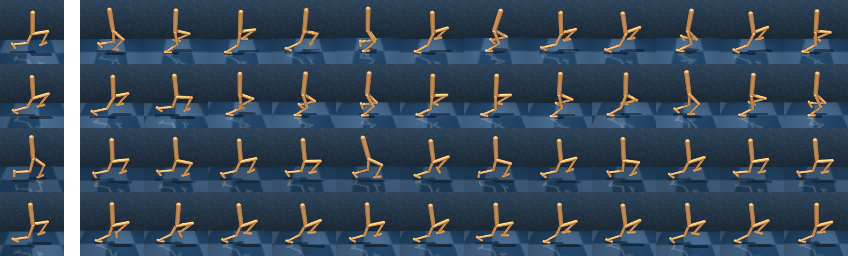}
        \caption{$32$ length.}
    \end{subfigure}
    \begin{subfigure}[b]{\textwidth}
        \centering
        \includegraphics[width=\textwidth]{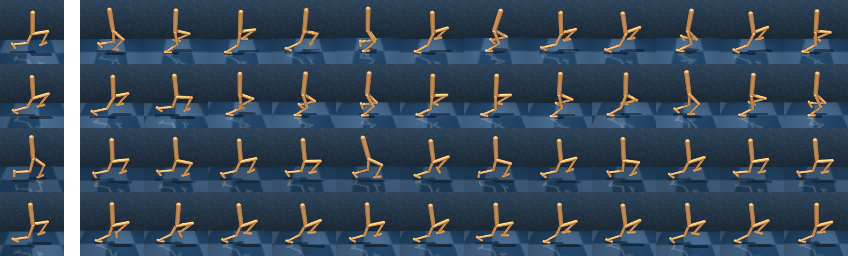}
        \caption{$16$ length.}
    \end{subfigure}
    \begin{subfigure}[b]{\textwidth}
        \centering
        \includegraphics[width=\textwidth]{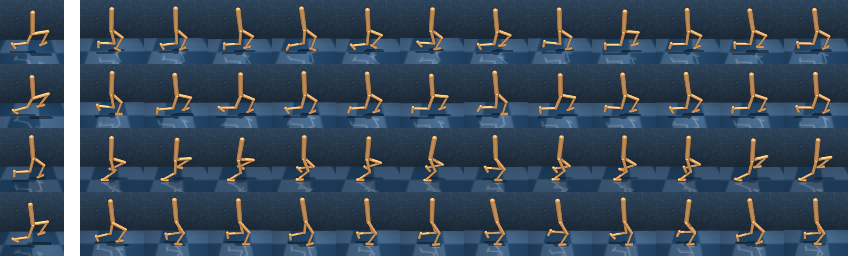}
        \caption{$8$ length.}
    \end{subfigure}
    \caption{Sample goals from the \texttt{walker\_run} task learned using the external and the exploratory rewards. The images on the left show the current state $s_t$, and the remaining images show the goal options generated by different skill CVAEs.}
    \label{fig:walkerrun_skill_goal_samples}
\end{figure}

\begin{figure}[h]
    \centering
    \begin{subfigure}[b]{\textwidth}
        \centering
        \includegraphics[width=\textwidth]{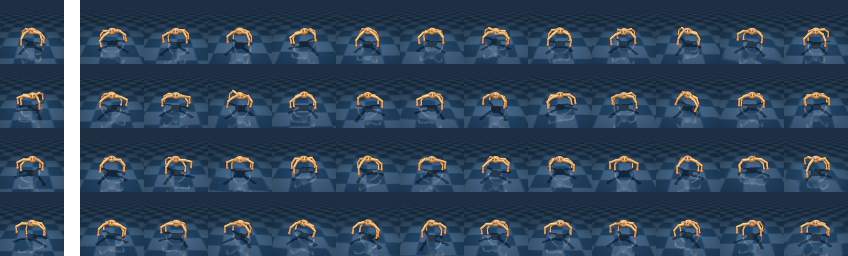}
        \caption{$64$ length.}
    \end{subfigure}
    \begin{subfigure}[b]{\textwidth}
        \centering
        \includegraphics[width=\textwidth]{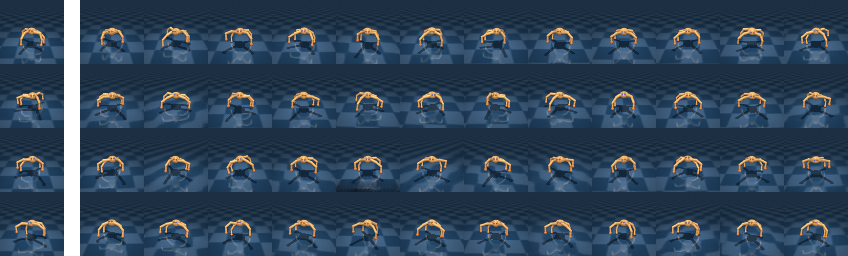}
        \caption{$32$ length.}
    \end{subfigure}
    \begin{subfigure}[b]{\textwidth}
        \centering
        \includegraphics[width=\textwidth]{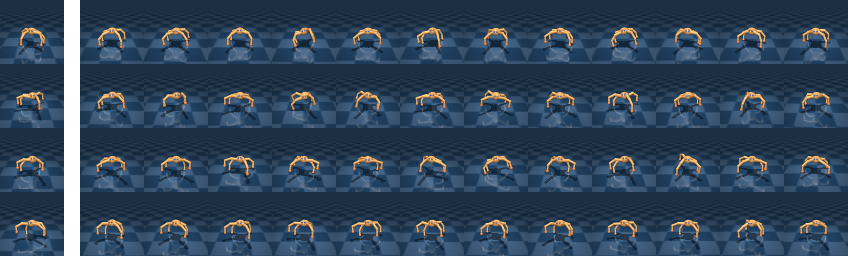}
        \caption{$16$ length.}
    \end{subfigure}
    \begin{subfigure}[b]{\textwidth}
        \centering
        \includegraphics[width=\textwidth]{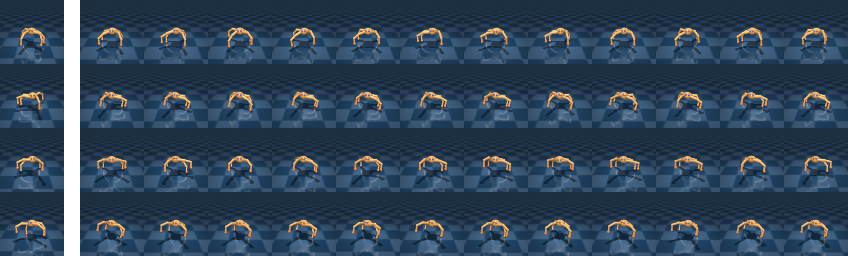}
        \caption{$8$ length.}
    \end{subfigure}
    \caption{Sample goals from the \texttt{quadruped\_run} task.}
    \label{fig:quadrun_skill_goal_samples}
\end{figure}

\begin{figure}[h]
    \centering
    \begin{subfigure}[b]{\textwidth}
        \centering
        \includegraphics[width=\textwidth]{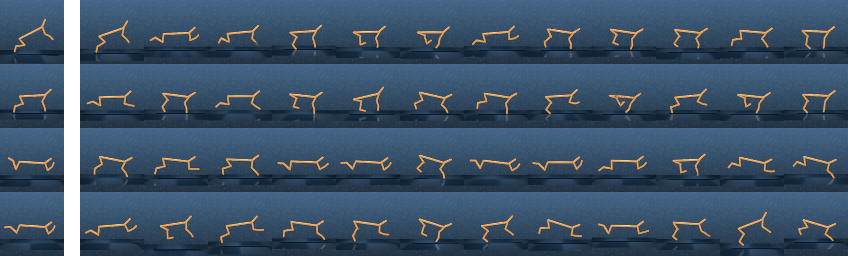}
        \caption{$64$ length.}
    \end{subfigure}
    \begin{subfigure}[b]{\textwidth}
        \centering
        \includegraphics[width=\textwidth]{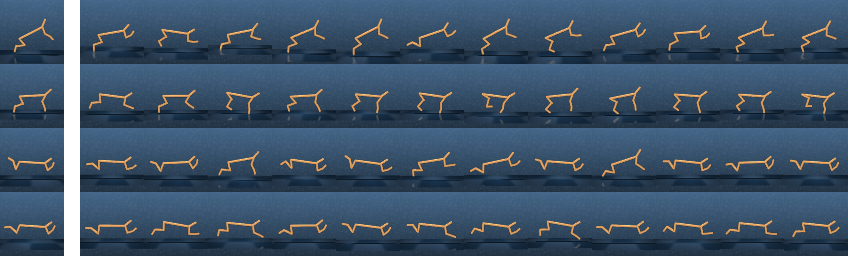}
        \caption{$32$ length.}
    \end{subfigure}
    \begin{subfigure}[b]{\textwidth}
        \centering
        \includegraphics[width=\textwidth]{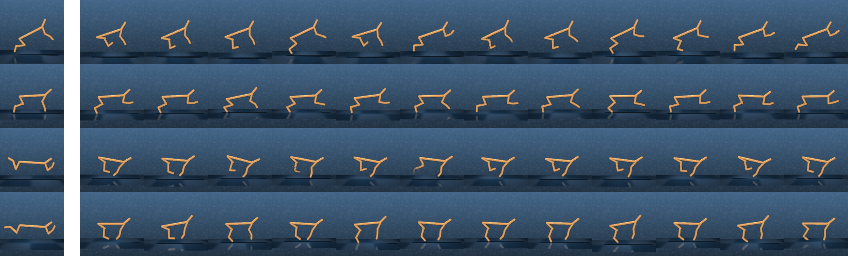}
        \caption{$16$ length.}
    \end{subfigure}
    \begin{subfigure}[b]{\textwidth}
        \centering
        \includegraphics[width=\textwidth]{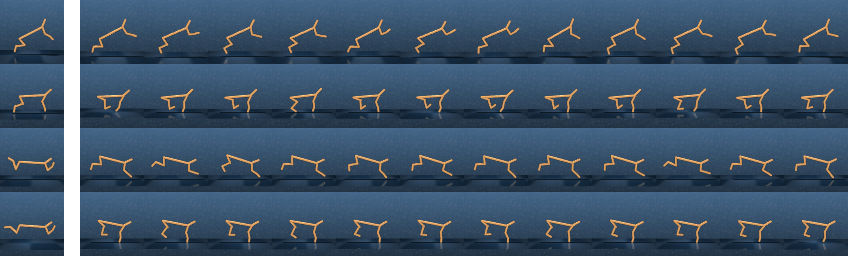}
        \caption{$8$ length.}
    \end{subfigure}
    \caption{Sample goals from the \texttt{cheetah\_run} task.}
    \label{fig:cheetahrun_skill_goal_samples}
\end{figure}

\begin{figure}[h]
    \centering
    \begin{subfigure}[b]{\textwidth}
        \centering
        \includegraphics[width=\textwidth]{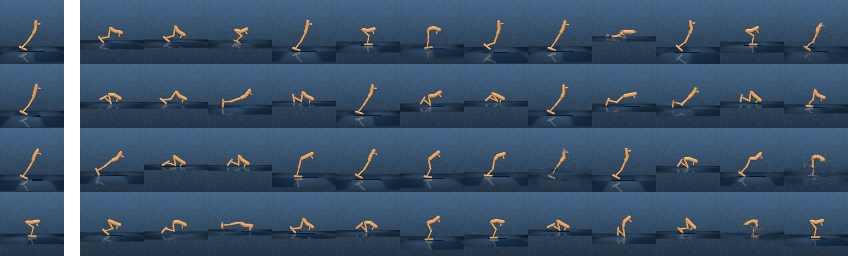}
        \caption{$64$ length.}
    \end{subfigure}
    \begin{subfigure}[b]{\textwidth}
        \centering
        \includegraphics[width=\textwidth]{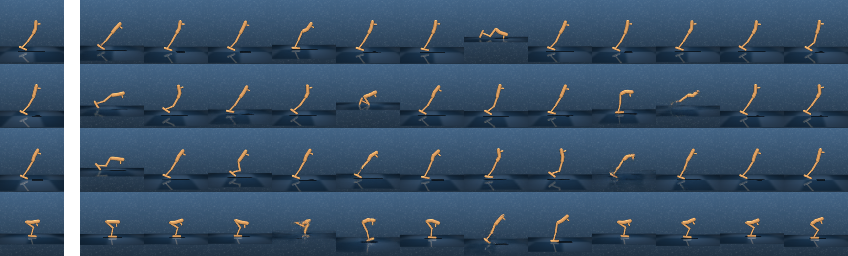}
        \caption{$32$ length.}
    \end{subfigure}
    \begin{subfigure}[b]{\textwidth}
        \centering
        \includegraphics[width=\textwidth]{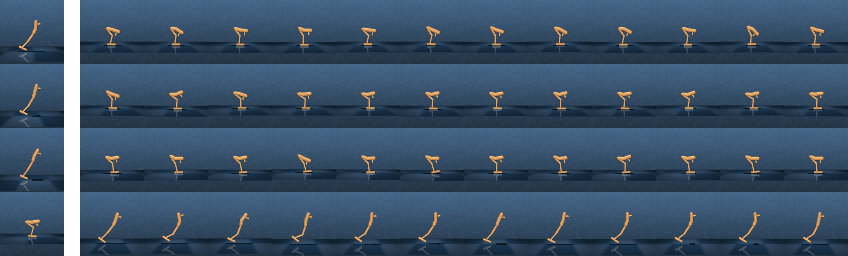}
        \caption{$16$ length.}
    \end{subfigure}
    \begin{subfigure}[b]{\textwidth}
        \centering
        \includegraphics[width=\textwidth]{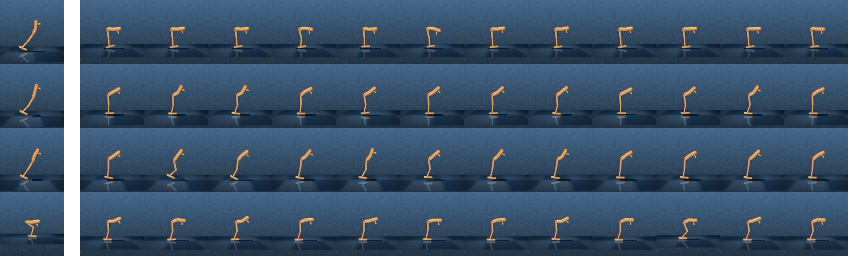}
        \caption{$8$ length.}
    \end{subfigure}
    \caption{Sample goals from the \texttt{hopper\_hop} task.}
    \label{fig:hopperhop_skill_goal_samples}
\end{figure}

The agent is trained using the default objective (weighted external and exploratory advantages).
Since we use a strong bias towards the external reward $([1.0,0.1])$, the learned skills are biased toward goal states that are more appropriate for the objective.
We sample the goals for tasks: \texttt{walker\_run} (Fig. \ref{fig:walkerrun_skill_goal_samples}), \texttt{quadruped\_run} (Fig. \ref{fig:quadrun_skill_goal_samples}), \texttt{cheetah\_run} (Fig. \ref{fig:cheetahrun_skill_goal_samples}), and \texttt{hopper\_hop} (Fig. \ref{fig:hopperhop_skill_goal_samples}).

\clearpage

\subsection{Exploration Only}

\begin{figure}[h]
    \centering
    \begin{subfigure}[b]{\textwidth}
        \centering
        \includegraphics[width=0.8\textwidth]{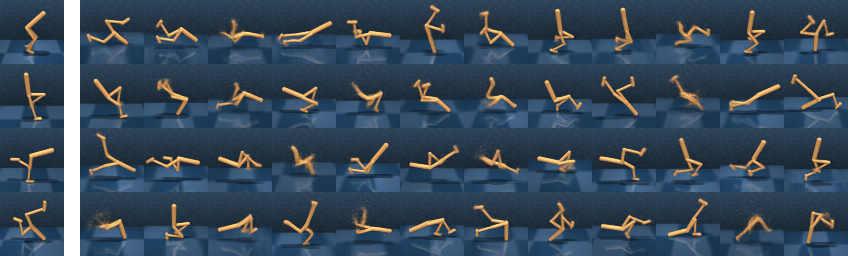}
        \caption{$64$ length.}
    \end{subfigure}
    \begin{subfigure}[b]{\textwidth}
        \centering
        \includegraphics[width=0.8\textwidth]{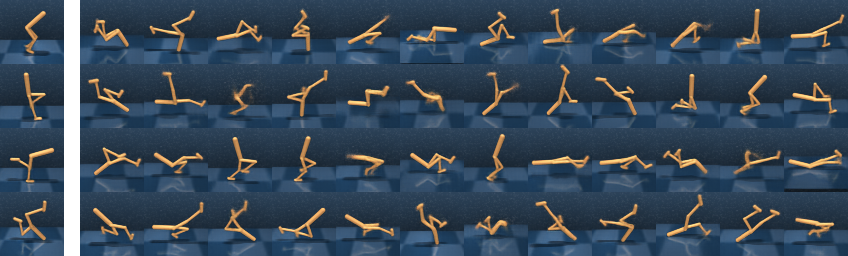}
        \caption{$32$ length.}
    \end{subfigure}
    \begin{subfigure}[b]{\textwidth}
        \centering
        \includegraphics[width=0.8\textwidth]{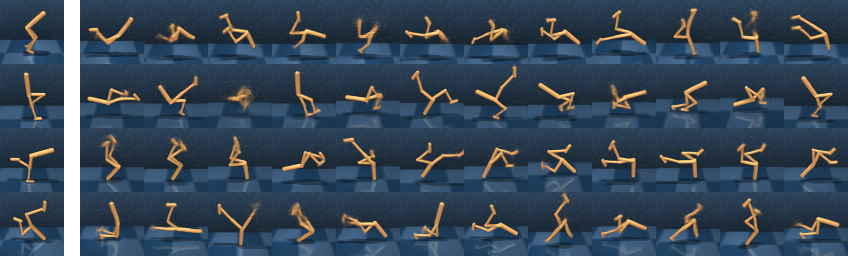}
        \caption{$16$ length.}
    \end{subfigure}
    \begin{subfigure}[b]{\textwidth}
        \centering
        \includegraphics[width=0.8\textwidth]{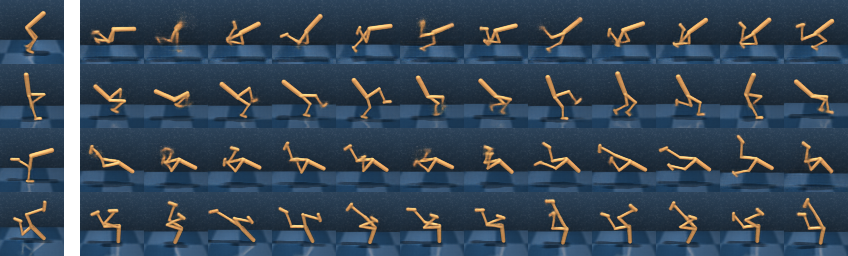}
        \caption{$8$ length.}
    \end{subfigure}
    \caption{Sample goals learned using only the exploratory objective in a \texttt{walker} embodiment. The images on the left show the current state $s_t$, and the remaining images show the goal options generated by different skill CVAEs.}
    \label{fig:walker_expl_goal_samples}
\end{figure}

\begin{figure}[h]
    \centering
    \begin{subfigure}[b]{\textwidth}
        \centering
        \includegraphics[width=\textwidth]{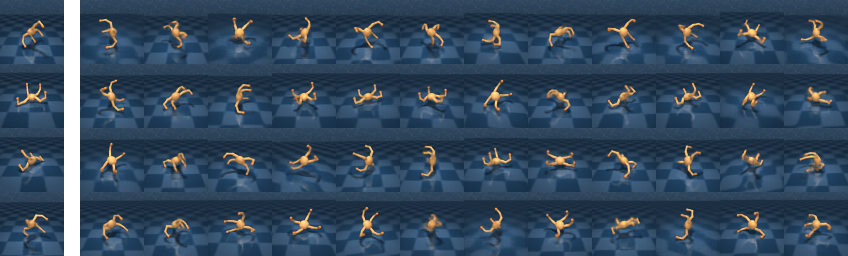}
        \caption{$64$ length.}
    \end{subfigure}
    \begin{subfigure}[b]{\textwidth}
        \centering
        \includegraphics[width=\textwidth]{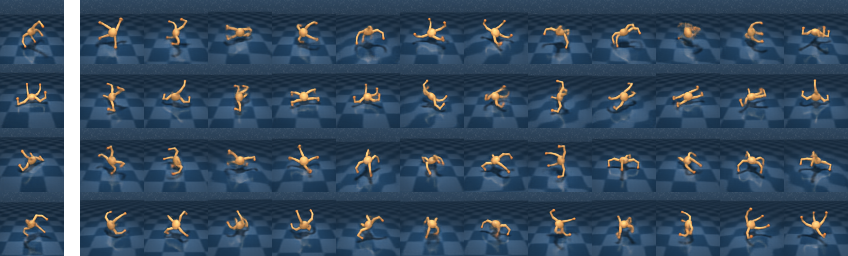}
        \caption{$32$ length.}
    \end{subfigure}
    \begin{subfigure}[b]{\textwidth}
        \centering
        \includegraphics[width=\textwidth]{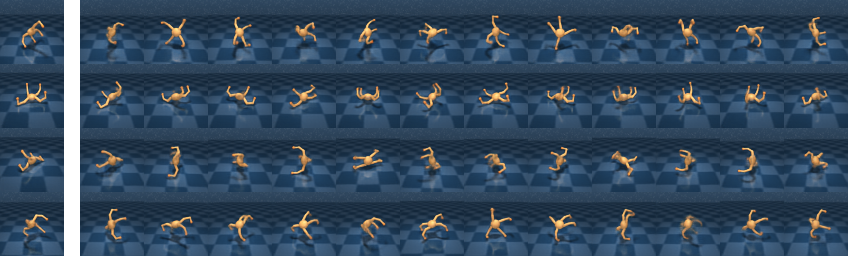}
        \caption{$16$ length.}
    \end{subfigure}
    \begin{subfigure}[b]{\textwidth}
        \centering
        \includegraphics[width=\textwidth]{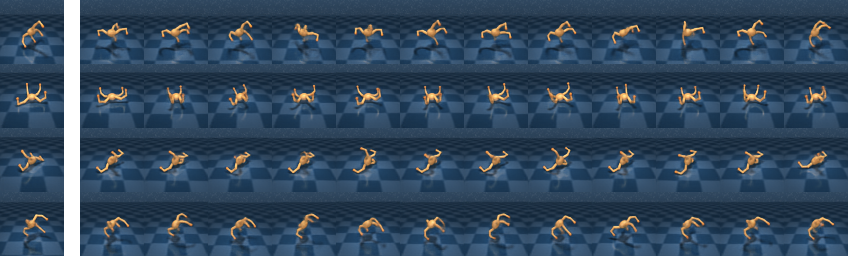}
        \caption{$8$ length.}
    \end{subfigure}
    \caption{Sample goals from exploration as a \texttt{quadruped}.}
    \label{fig:quad_expl_goal_samples}
\end{figure}

\begin{figure}[h]
    \centering
    \begin{subfigure}[b]{\textwidth}
        \centering
        \includegraphics[width=\textwidth]{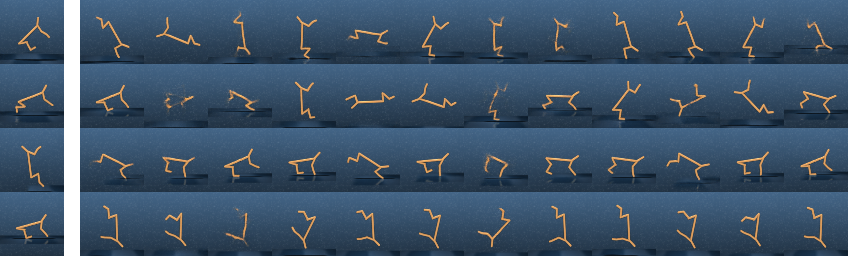}
        \caption{$64$ length.}
    \end{subfigure}
    \begin{subfigure}[b]{\textwidth}
        \centering
        \includegraphics[width=\textwidth]{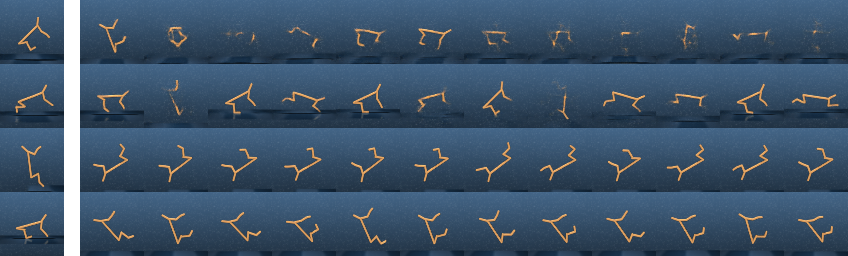}
        \caption{$32$ length.}
    \end{subfigure}
    \begin{subfigure}[b]{\textwidth}
        \centering
        \includegraphics[width=\textwidth]{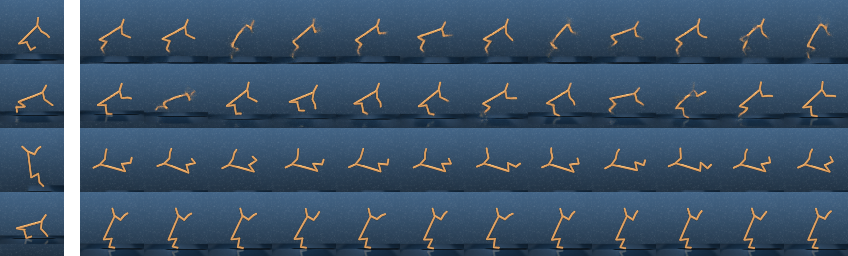}
        \caption{$16$ length.}
    \end{subfigure}
    \begin{subfigure}[b]{\textwidth}
        \centering
        \includegraphics[width=\textwidth]{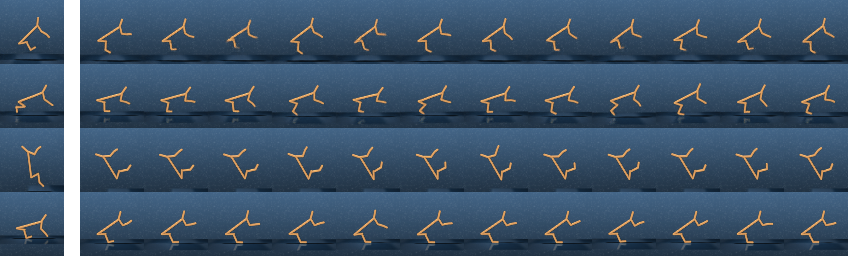}
        \caption{$8$ length.}
    \end{subfigure}
    \caption{Sample goals from exploration as a \texttt{cheetah}.}
    \label{fig:cheetah_expl_goal_samples}
\end{figure}

\begin{figure}[h]
    \centering
    \begin{subfigure}[b]{\textwidth}
        \centering
        \includegraphics[width=\textwidth]{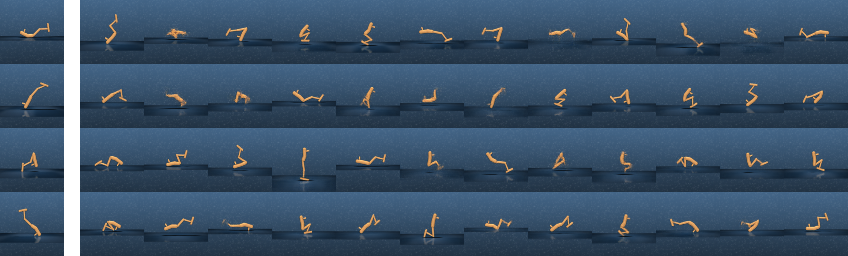}
        \caption{$64$ length.}
    \end{subfigure}
    \begin{subfigure}[b]{\textwidth}
        \centering
        \includegraphics[width=\textwidth]{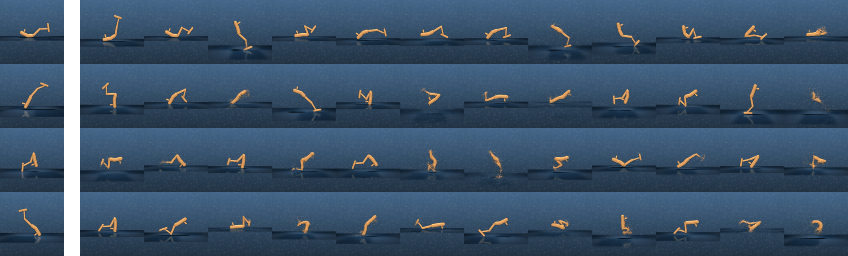}
        \caption{$32$ length.}
    \end{subfigure}
    \begin{subfigure}[b]{\textwidth}
    
        \centering
        \includegraphics[width=\textwidth]{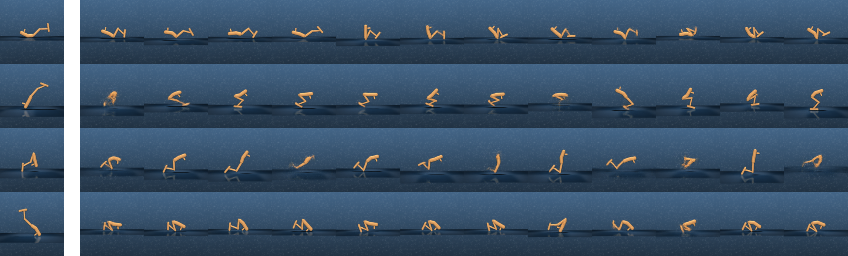}
        \caption{$16$ length.}
    \end{subfigure}
    \begin{subfigure}[b]{\textwidth}
        \centering
        \includegraphics[width=\textwidth]{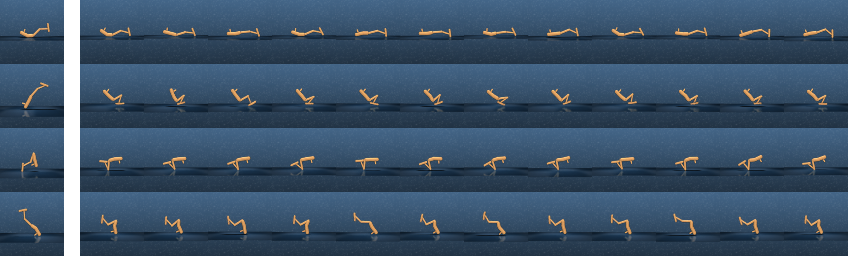}
        \caption{$8$ length.}
    \end{subfigure}
    \caption{Sample goals from exploration as a \texttt{Hopper}.}
    \label{fig:hopper_expl_goal_samples}
\end{figure}

The agent is optimized solely for the exploration objective, which aims to maximize coverage of the state-transition space.
We sample the goals per Skill CVAE for embodiments: \texttt{walker} (Fig. \ref{fig:walker_expl_goal_samples}), \texttt{quadruped} (Fig. \ref{fig:quad_expl_goal_samples}), \texttt{cheetah} (Fig. \ref{fig:cheetah_expl_goal_samples}), and \texttt{hopper} (Fig. \ref{fig:hopper_expl_goal_samples}) in the DMC suite \cite{tassa2018deepmind}.

\clearpage

\section{Visualizing Choice Preferences for States}
\label{sec:states_per_choice}

\begin{figure}[h]
    \centering
    \begin{subfigure}[b]{\textwidth}
        \centering
        \includegraphics[width=\textwidth]{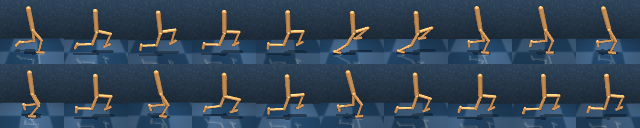}
        \caption{$64$ length.}
    \end{subfigure}
    \begin{subfigure}[b]{\textwidth}
        \centering
        \includegraphics[width=\textwidth]{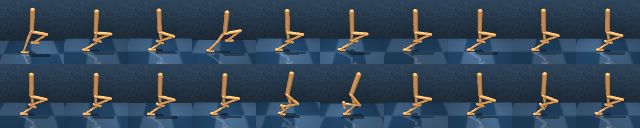}
        \caption{$32$ length.}
    \end{subfigure}
    \begin{subfigure}[b]{\textwidth}
        \centering
        \includegraphics[width=\textwidth]{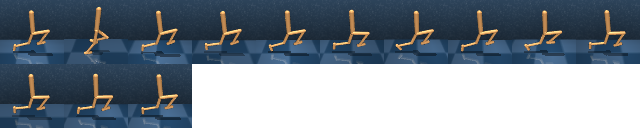}
        \caption{$16$ length.}
    \end{subfigure}
    \begin{subfigure}[b]{\textwidth}
        \centering
        \includegraphics[width=\textwidth]{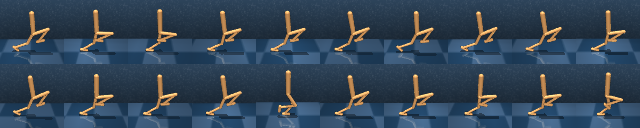}
        \caption{$8$ length.}
    \end{subfigure}
    \begin{subfigure}[b]{\textwidth}
        \centering
        \includegraphics[width=\textwidth]{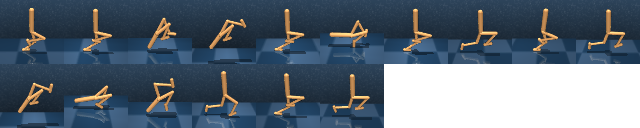}
        \caption{$\infty$ length.}
    \end{subfigure}
    \caption{States segregated by choice for the \texttt{walker\_run} task.}
    \label{fig:walker_run_choice_state_samples}
\end{figure}

\begin{figure}[h]
    \centering
    \begin{subfigure}[b]{\textwidth}
        \centering
        \includegraphics[width=\textwidth]{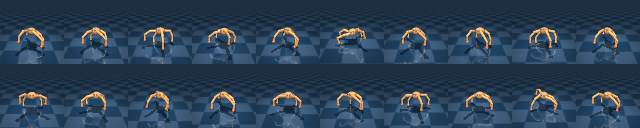}
        \caption{$64$ length.}
    \end{subfigure}
    \begin{subfigure}[b]{\textwidth}
        \centering
        \includegraphics[width=\textwidth]{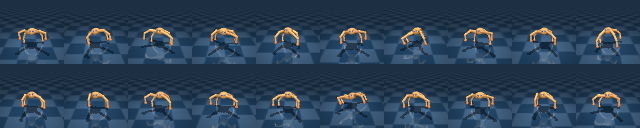}
        \caption{$32$ length.}
    \end{subfigure}
    \begin{subfigure}[b]{\textwidth}
        \centering
        \includegraphics[width=\textwidth]{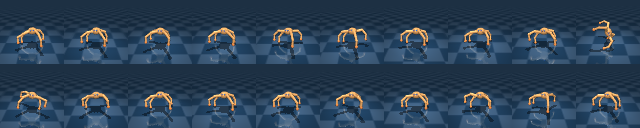}
        \caption{$16$ length.}
    \end{subfigure}
    \begin{subfigure}[b]{\textwidth}
        \centering
        \includegraphics[width=\textwidth]{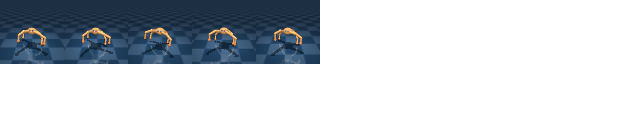}
        \caption{$8$ length.}
    \end{subfigure}
    \begin{subfigure}[b]{\textwidth}
        \centering
        \includegraphics[width=\textwidth]{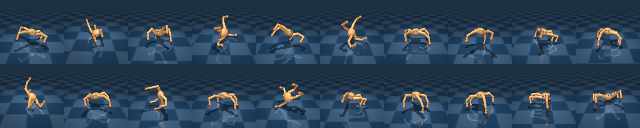}
        \caption{$\infty$ length.}
    \end{subfigure}
    \caption{States segregated by choice for the \texttt{quadruped\_run} task.}
    \label{fig:quadruped_run_choice_state_samples}
\end{figure}

\begin{figure}[h]
    \centering
    \begin{subfigure}[b]{\textwidth}
        \centering
        \includegraphics[width=\textwidth]{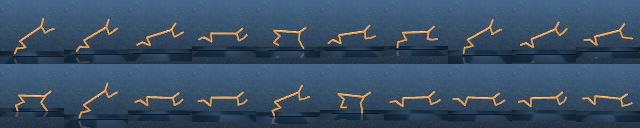}
        \caption{$64$ length.}
    \end{subfigure}
    \begin{subfigure}[b]{\textwidth}
        \centering
        \includegraphics[width=\textwidth]{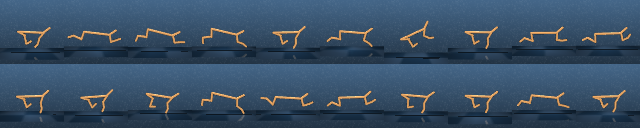}
        \caption{$32$ length.}
    \end{subfigure}
    \begin{subfigure}[b]{\textwidth}
        \centering
        \includegraphics[width=\textwidth]{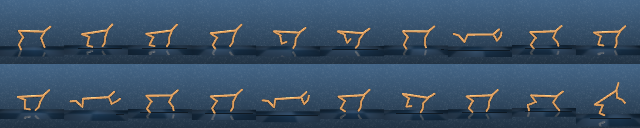}
        \caption{$16$ length.}
    \end{subfigure}
    \begin{subfigure}[b]{\textwidth}
        \centering
        \includegraphics[width=\textwidth]{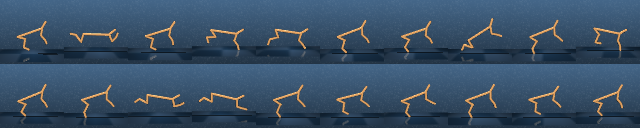}
        \caption{$8$ length.}
    \end{subfigure}
    \begin{subfigure}[b]{\textwidth}
        \centering
        \includegraphics[width=\textwidth]{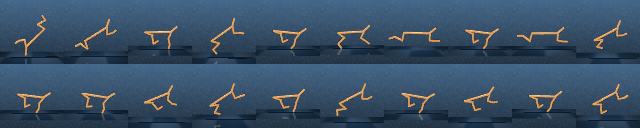}
        \caption{$\infty$ length.}
    \end{subfigure}
    \caption{States segregated by choice for the \texttt{cheetah\_run} task.}
    \label{fig:cheetah_run_choice_state_samples}
\end{figure}

\begin{figure}[h]
    \centering
    \begin{subfigure}[b]{\textwidth}
        \centering
        \includegraphics[width=\textwidth]{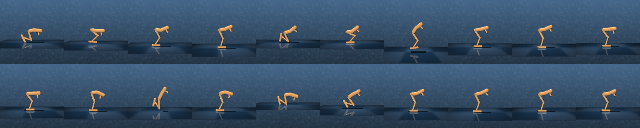}
        \caption{$64$ length.}
    \end{subfigure}
    \begin{subfigure}[b]{\textwidth}
        \centering
        \includegraphics[width=\textwidth]{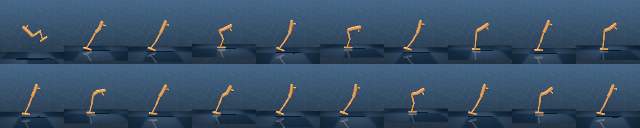}
        \caption{$32$ length.}
    \end{subfigure}
    \begin{subfigure}[b]{\textwidth}
        \centering
        \includegraphics[width=\textwidth]{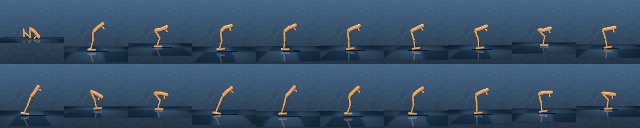}
        \caption{$16$ length.}
    \end{subfigure}
    \begin{subfigure}[b]{\textwidth}
        \centering
        \includegraphics[width=\textwidth]{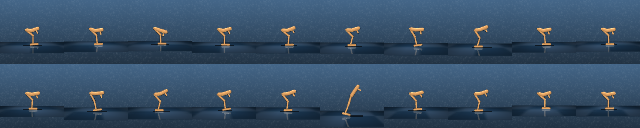}
        \caption{$8$ length.}
    \end{subfigure}
    \begin{subfigure}[b]{\textwidth}
        \centering
        \includegraphics[width=\textwidth]{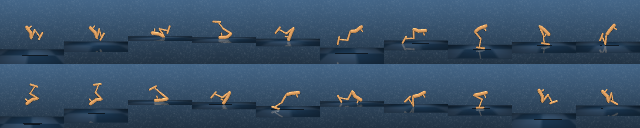}
        \caption{$\infty$ length.}
    \end{subfigure}
    \caption{States segregated by choice for the \texttt{hopper\_hop} task.}
    \label{fig:hopper_hop_choice_state_samples}
\end{figure}

To verify if the agent exhibits any skill-length preferences in states, we run the agent for $5$ episodes.
Then the states $s_t$ visited by the agent are segregated by the choice ($c_t$) made by the agent and shown below.
The visualizations can help verify if there are any correlations between the agent state $s_t$ and the choice variable $s_t$.
We sample the states for the tasks: \texttt{walker\_run} (Fig. \ref{fig:walker_run_choice_state_samples}), \texttt{quadruped\_run} (Fig. \ref{fig:quadruped_run_choice_state_samples}), \texttt{cheetah\_run} (Fig. \ref{fig:cheetah_run_choice_state_samples}), and \texttt{hopper\_hop} (Fig. \ref{fig:hopper_hop_choice_state_samples}).
A common trend observed across tasks is that $\inf$-skills are preferred in uncommon states, such as upside-down or recovering after a misstep.

\end{document}